\documentclass[review]{elsarticle}
\usepackage{lineno,hyperref}
\hypersetup{colorlinks = true}
\modulolinenumbers[5]
\usepackage{bm}
\usepackage{color}

\usepackage{graphicx, subcaption}
\usepackage{multirow}
\usepackage{makecell}
\usepackage{amsmath}
\usepackage{algpseudocode}
\usepackage{algorithm}
\algnewcommand\algorithmicinput{\textbf{Input:}}
\algnewcommand\Input{\item[\algorithmicinput]}
\usepackage{tabularx}
\usepackage{amssymb}
\usepackage{threeparttable}
\newcolumntype{Y}{>{\centering\arraybackslash}X}
\usepackage{placeins}
\graphicspath{{./Figures/}}

\newcommand{\hlight}[1]{{\color{black} #1}}
\DeclareMathOperator*{\argmax}{argmax}
\DeclareMathOperator*{\argmin}{argmin}

\journal{Engineering Applications of Artificial Intelligence}

 \usepackage{multirow}
\usepackage[table,xcdraw]{xcolor}

\usepackage[some]{background}
\SetBgScale{1}
\SetBgContents{\parbox{10cm}{%
  \Huge Draft:  \today\\[14cm]\rotatebox{180}{\Huge Draft:  \today}}}
\SetBgColor{red}
\SetBgAngle{270}
\SetBgOpacity{0.2}

\biboptions{authoryear}

\begin{document}

\begin{frontmatter}


\title{\hlight{Lazy FSCA for Unsupervised Variable Selection}}
%
\author[FirstAffiliation]{Federico Zocco}
\author[SecondAffiliation]{Marco Maggipinto}
\author[SecondAffiliation]{Gian Antonio Susto}
\author[FirstAffiliation]{Se\'{a}n McLoone}

\address[FirstAffiliation]{Centre for Intelligent Autonomous Manufacturing Systems (i-AMS), Queen's University Belfast, Northern Ireland, UK; email: \{f.zocco,s.mcloone\}@qub.ac.uk}
\address[SecondAffiliation]{Department of Information Engineering, University of Padua, Italy; \\ email: marco.maggipinto@gmail.com, gianantonio.susto@dei.unipd.it}

\begin{abstract}

\hlight{Various unsupervised greedy selection methods have been proposed as computationally tractable approximations to the NP-hard subset selection problem. These methods rely on sequentially selecting the variables that best improve performance with respect to a selection criterion. Theoretical results exist that provide performance bounds and enable ‘lazy greedy’ efficient implementations for selection criteria that satisfy a diminishing returns property known as submodularity. This has motivated the development of variable selection algorithms based on mutual information and frame potential.  Recently, the authors introduced Forward Selection Component Analysis (FSCA) which uses variance explained as its selection criterion. While this criterion is not submodular, FSCA has been shown to be highly effective for applications such as measurement plan optimisation.  In this paper a ‘lazy’ implementation of the FSCA algorithm (L-FSCA) is proposed, which, although not equivalent to FSCA due to the absence of submodularity, has the potential to yield comparable performance while being up to an order of magnitude faster to compute. The efficacy of L-FSCA is demonstrated by performing a systematic comparison with FSCA and five other unsupervised variable selection methods from the literature using simulated and real-world case studies. Experimental results confirm that L-FSCA yields almost identical performance to FSCA while reducing computation time by between 22\% and 94\% for the case studies considered.
} 
  
\end{abstract}


\end{frontmatter}

\section{\hlight{Introduction}} \label{intro}
%
%

When dealing with high dimensional datasets, unsupervised dimensionality reduction is often performed as a pre-processing step to achieve efficient storage of the data and robustness of machine learning algorithms. The underlying assumption is that such datasets
have high levels of correlation among variables and hence redundancy that can
be exploited. Principal Component Analysis (PCA) is the most popular linear technique for unsupervised dimensionality reduction \citep{jolliffe1986principal,Maaten2009}, consisting of the transformation of variables into a set of orthogonal components that correspond to the directions of maximum variance in the data.    

In some applications it is desirable to achieve unsupervised dimensionality reduction through the selection of a subset of initial variables that best represent the information contained in the full dataset, for example in sensor selection problems \citep{krause2008near,mcloone2018methodology}. Working with a subset of the original variables also facilitates more transparent and interpretable models than those based on principal components \citep{flynn2011max}. The unsupervised variable selection problem may be formulated as follows. Given a dataset $\mathbf{X} \in \mathbb{R}^{m \times v}$ containing $m$ measurements of $v$ variables, and the index set for the variables (i.e. columns of $\mathbf{X}$) $I_X = \{1, 2, \dots, v-1, v\}$ and defining $I_S$ and $I_U$ as the indices of the selected and unselected variables ($I_S \cup I_U = I_X$), respectively, we wish to solve
\begin{equation}
\label{eq:VSProblem01}
I_S^* = \underset{I_S \subset I_X}{\argmax} \,g(I_S)  ~~~ s.t. ~~ |I_S|=k,
\end{equation}
where $I_S^*$ denotes the optimum subset, $k$ is a cardinality constraint on  $I_S$  and $g(\cdot)$: $2^{v} \mapsto \mathbb{R}$ is a performance metric (set function) which measures the suitability of the selected variables. An alternative formulation of the variable selection problem places a constraint $\tau$ on the target value for the performance metric, that is:
\begin{equation}
\label{eq:VSProblem02}
I_S^* = \underset{I_S \subset I_X}{\argmin} \, |I_S| ~~~ s.t. ~~ g(I_S) > \tau.
\end{equation}

\begin{figure}
	\centering
	\includegraphics[width=\textwidth]{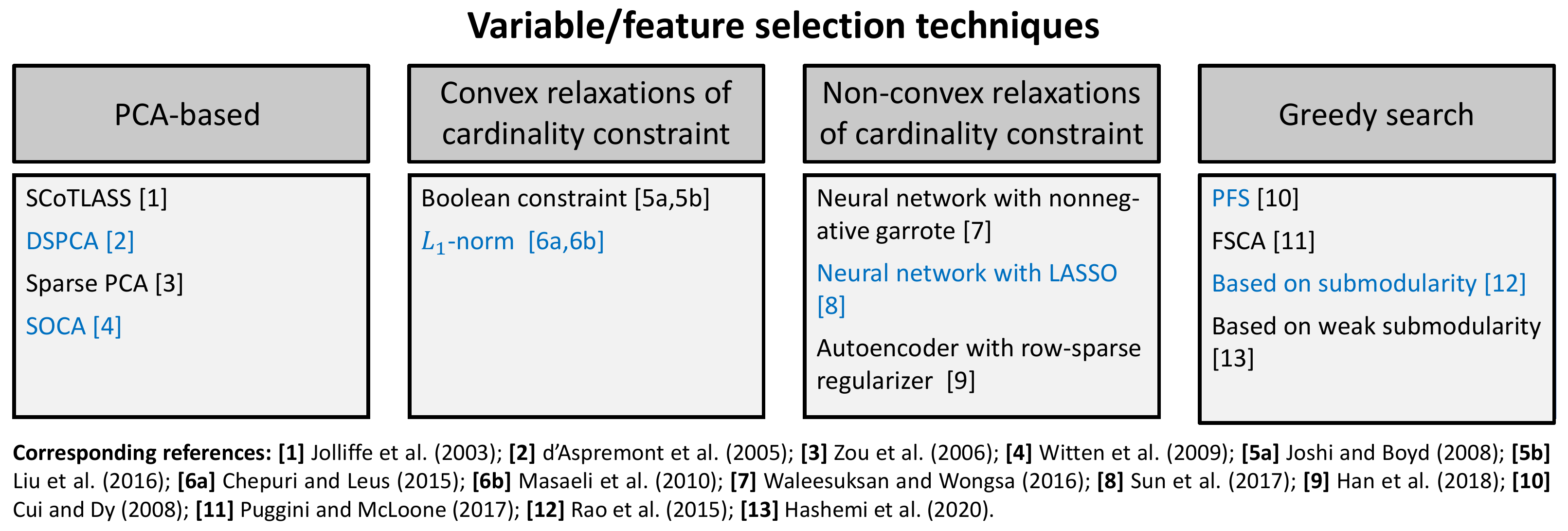} 
	\caption{\hlight{A taxonomy of approximate solution methods (with examples) for the NP-hard variable/feature selection problem.}}
	\label{fig:IntroductionScheme}
\end{figure}

Finding the optimal subset of variables from a set of candidate variables is an NP-hard combinatorial problem and therefore intractable in practice even for relatively low dimension problems. Consequently, developing computationally efficient techniques that approximate the optimum solution have been the focus of research for many years. Existing techniques can be split into four categories as depicted in \hlight{Fig. \ref{fig:IntroductionScheme}.}
%
%
\nocite{jolliffe2003modified,d2005direct,zou2006sparse,witten2009penalized} 
\nocite{joshi2008sensor,liu2016sensor,chepuri2015sparsity,masaeli2010convex}
\nocite{waleesuksan2016fast,sun2017design,Han2018} 
\nocite{cui2008orthogonal,IEEEPuggini,rao2015greedy,hashemi2020}%
Forward greedy search methods, which are the focus of this study, estimate the optimum subset of variables by recursively adding them one at a time, such that the variable added at each iteration is the one that gives the optimal improvement in the chosen performance metric $g(\cdot)$. The general structure of a forward greedy selection algorithm for Eq. (\ref{eq:VSProblem01}) is as defined in Algorithm \ref{alg:general}. The corresponding greedy algorithm for solving Eq. (\ref{eq:VSProblem02}) can be obtained by replacing the while loop condition in line 2 with $g(I_S) < \tau$.

\begin{algorithm}[!h]
	\caption{Forward greedy variable selection}
	\begin{algorithmic}[1] 
		\Input $\bm{X}, k$
		\State $I_U=\{1, 2, \dots, v-1, v\};~ I_S = \emptyset$
		\While {$|I_S|<k$}
		\State $i^{*} = \underset{i \in I_U}{\argmax} \, \{g(I_S \cup \{i\})\}$
		\State $I_S = I_S \cup \{i^*\}; I_U = I_U\setminus \{i^*\}$
		\EndWhile \\
		\Return $I_S$
	\end{algorithmic}
	\label{alg:general}
\end{algorithm}

In general, the outcome of a greedy search has no optimality guarantees. However, if $g(\cdot)$ is a \emph{monotone increasing}, \emph{submodular}\footnote{Formally defined in Section \ref{Background_Theory}.} metric and $g(\emptyset) = 0$, then the greedy search solution $g(I_S)$ is guaranteed to be within $63\%$ of the optimal solution and a more efficient implementation of the greedy search algorithm, referred to as the \emph{lazy} greedy implementation \citep{minoux1978}, is possible. 

Several greedy search based unsupervised variable selection algorithms have been proposed in the literature using a number of different performance metrics. \emph{Variance explained} (VE) was employed in the forward selection component analysis (FSCA) algorithm introduced in \cite{IEEEPuggini} and \cite{Prakash2012a}, while the squared multiple correlation was employed by \cite{wei2007feature} in their forward orthogonal selection maximising overall dependency (FOS-MOD) algorithm. \hlight{VE also underpins the orthogonal principal feature selection (PFS) algorithm proposed in \cite{cui2008orthogonal}. In PFS variables are selected based on their correlation with the first principal component of the subspace orthogonal to that spanned by the currently selected variables. In contrast,} the unsupervised forward selection (UFS) algorithm by \cite{whitley2000unsupervised} selects the variables having the lowest squared multiple correlation with the subspace spanned by the currently selected variables. While these metrics are natural choices for unsupervised variable selection given their close relationship with linear regression and PCA, they are not submodular functions. Motivated by the optimality bound and more efficient algorithm implementation that follow from submodularity, \cite{krause2008near} used \emph{mutual information} (MI), \cite{ranieri2014near} used \emph{frame potential} (FP) and  \cite{joshi2008sensor} and \cite{rao2015greedy} used \emph{log-det} as submodular metrics to design near-optimal greedy sensor selection algorithms. 

To date the literature lacks a comparison of the main unsupervised greedy selection algorithms. There is also a lack of awareness of the benefit of using a lazy greedy implementation even when submodularity does not apply. While VE does not satisfy the conditions for submodularity, and therefore does not enjoy convenient to compute theoretical bounds on its performance, practical experience shows that greedy search algorithms based on VE perform very well. This may mean that for many problems VE is close to being submodular, and indeed, as observed by \cite{das2008algorithms}, it is provably so in certain cases.

Motivated by these observations, this paper makes the following \hlight{novel} contributions. \hlight{
\begin{itemize}
\item{A lazy greedy implementation of FSCA (denoted L-FSCA) is proposed that is able to achieve comparable performance to FSCA while being up to an order of magnitude faster to compute.} 

\item{L-FSCA and the main unsupervised greedy variable selection algorithms proposed in the literature are formalised within a common framework and compared with respect to Big-O computational complexity.}
	
\item{The algorithms are then evaluated experimentally with respect to the aforementioned performance metrics (VE, MI and FP) on a diverse set of simulated and real world benchmark datasets. While demonstrating the efficacy of L-FSCA, this also serves as the first systematic comparison of unsupervised variable selection algorithms in the literature.}



\end{itemize}
}
\noindent In addition, to facilitate benchmarking against other unsupervised variable selection  algorithms in the future, the code for the algorithms and experimental studies presented in the paper have been made available on GitHub\footnote{\url{https://github.com/fedezocco/GreedyVarSel-MATLAB}}.

The remainder of the paper is organised as follows: 
\hlight{Section \ref{Background_Theory} provides a summary of the main theoretical results on performance guarantees that apply to the unsupervised variable selection problem at hand, and introduces the lazy greedy algorithm that can be exploited when selection metrics are submodular.
 Section \ref{meth} then briefly introduces the unsupervised variable selection algorithms and performance metrics under investigation and provides an analysis of their computational complexity.} The performance comparison on simulated and real datasets is presented in Section \ref{res} and conclusions are provided in Section \ref{concl}. The following notation is adopted: matrices and vectors are indicated with bold capital and lower case letters, respectively, while sets and scalars are denoted by capital and lower case letters, respectively.

\section {Underpinning Theory} \label{Background_Theory}
\subsection{Theoretical Guarantees on Performance} \label{sec: Performance bounds}
Let $v$, $k$, $I_S$ and $I^*_S$ denote the total number of candidate variables, the number of selected variables, the index set of the variables selected via greedy search and the corresponding optimal index set, respectively. 
If the greedy unsupervised variable selection metric $g(\cdot)$ is a monotone increasing, submodular set function and $g(\emptyset) = 0$, then, according to the seminal result by \cite{nemhauser1978analysis},  the greedy search solution $g(I_S)$ is guaranteed to be within $63\%$ of the optimal solution, or more specifically:  
\begin{equation}
\mathcal{B}_{\text{N}}:~~~	\frac{g(I_S)}{g(I^*_S)} \geq 1- \left(\frac{k-1}{k}\right)^k \geq 1-1/e \geq 0.63.
	\label{eq:SubModularOptBound}
	\end{equation}
A set function $g(\cdot)$ is defined as being \emph{monotone increasing} if $\forall A \subseteq B \subseteq X$, it holds that $g(A) \leq g(B)$. It is defined as \emph{submodular}  \citep{das2008algorithms} if  $\forall A \subseteq  B \subseteq X, \forall x \in X\setminus B$ it satisfies the property
\begin{equation}
 g(A \cup \{x\}) - g(A) \geq g(B \cup \{x\}) - g(B).
 \label{eq:SubmodularityDef}
\end{equation}
If the condition in Eq. (\ref{eq:SubmodularityDef}) holds with equality then $g(\cdot)$ is a \emph{modular} function and the greedy solution is guaranteed to be the optimal solution, i.e. $g(I_S) = g(I^*_S)$.  

In order to improve on $\mathcal{B}_{\text{N}}$ and obtain results for more general forms of set function the concepts of \emph{curvature} \citep{conforti1984,iyer2013curvature,sviridenko2017optimal} and \emph{submodularity ratio} \citep{das2011submodular, das2018approximate} have been introduced as measures of how far a submodular function is from being modular and how far a non-submodular function is from being submodular, respectively. \cite{bian2017guarantees} recently provided a unified treatment of these concepts and showed that if $g(\cdot)$ is a non-negative non-decreasing set function with curvature $\alpha \in [0,1]$ defined as
\begin{equation}
\alpha =  \max_{\substack{i,A,B \\{i \in X, A \subset B \subseteq X/i}}} \left[1 - \frac{g(B \cup \{i\}) - g(B)}{g(A \cup \{i\}) - g(A)} \right]
\label{eq:generalisedcurvature}
\end{equation}
and submodularity ratio $\gamma \in [0,1]$ defined as
\begin{equation}
\gamma = \min_{\substack{ A,B \\ A \subset X,B \subseteq X/A}} \left[\frac{\sum_{i \in B} \left(g(A \cup \{i\}) - g(A)\right)}{g(A \cup B) - g(A)}\right],
\label{eq:SubModularityRatio}
\end{equation}
then the following lower bound applies to the greedy solution
	\begin{equation}
\mathcal{B}_{\alpha\gamma}:~~~ \frac{g(I_S)}{g(I^*_S)} \geq \frac{1}{\alpha} \left[1-\left( \frac{k - \alpha \gamma} {k} \right)^k\right]  \geq  \frac{1}{\alpha}(1-e^{-\alpha \gamma}).
	\label{eq:NonSubModularOptBound}
	\end{equation}
When $\gamma$=1, $g(\cdot)$ is submodular. If $\gamma$=1 and $\alpha$=1,  $\mathcal{B}_{\alpha\gamma}$ yields Nemhauser et al.'s  bound (Eq. \ref{eq:SubModularOptBound}) and if $\alpha$=0, $g(\cdot)$ is \emph{supermodular} and $\mathcal{B}_{\alpha\gamma}$ reduces to
\begin{equation}
	\lim_{\alpha \rightarrow 0} \frac{1}{\alpha}(1-e^{-\alpha \gamma}) = \gamma.
	\label{eq:NonSubModularbound_sup}
\end{equation}
Finally, when $\alpha$=0 and $\gamma$=1, $g(\cdot)$ is modular and  $g(I_S) = g(I^*_S)$.

\hlight{Other definitions of curvature have also been introduced for non-submodular functions and used as a measure of how close a function is to being submodular, such as those by \cite{sviridenko2017optimal}, \cite{wang2016} and \cite{hashemi2019}, but the resulting bounds are weaker than $\mathcal{B}_{\alpha\gamma}$.}

\hlight{In general, computing these bounds is as computationally intractable as the original subset selection problem.  However, for specific problem formulations bounds can be computed for the curvature and submodularity ratio parameters that are computationally tractable, allowing weaker versions of the performance guarantees to be computed \citep{bian2017guarantees,das2011submodular}. Overall these bounds are highly conservative with greedy search algorithms frequently achieving near optimal solutions. As such, their primary value is providing a theoretical foundation for the effectiveness of greedy based search methods.}
 
It should be noted that even when $g(\cdot)$ is not a modular or submodular set function in general it may have these properties for restricted forms of $X$. For example, while variance explained (squared multiple correlation) is not a submodular function, \cite{das2008algorithms} show that if $X$ does not contain any suppressor variables, then $g(\cdot)$ will be submodular and Nemhauser et al.'s performance bound $\mathcal{B}_\text{N}$ applies.  Similarly, if $X$ is a set of uncorrelated variables, i.e $\mathbf{X}$ is an orthogonal matrix, then $g(\cdot)$ will be modular and the greedy search algorithm yields the optimal solution.

\subsection{\hlight{Submodularity and the Lazy Greedy Implementation}} 
Submodularity is essentially a diminishing returns property. Defining the \emph{marginal gain} (improvement in performance) of adding $x$ to $I_S$ at the $k$-th iteration of the greedy search algorithm as
\begin{equation}
\Delta g_k(x) = g(I_S \cup \{x\}) - g(I_S),
\end{equation}
Eq. (\ref{eq:SubmodularityDef}) implies that $\Delta g_j(x) \leq \Delta g_{k}(x)$ for $\forall j > k$, that is, the marginal gain achievable with a given variable decreases (or at best remains the same) the later in the sequence it is added. Consequently, the marginal gain given by an element, i.e. variable, $x$ at the current iteration is an upper bound on its marginal gain at subsequent iterations. This property can be exploited to arrive at a greedy search algorithm for submodular functions that has much lower computational complexity than the conventional greedy search algorithm, as first proposed by \cite{minoux1978}. The algorithm, usually referred to as the lazy greedy search algorithm, operates by maintaining a descending-ordered list of the upper bounds on the marginal gain of each candidate variable (which is initialized at the first iteration), and then, at each iteration, computing the marginal gains sequentially from the top of the list until a candidate variable is found whose \emph{marginal gain} is greater than the next largest \emph{upper bound} in the list. In this way, the number of set function evaluations required to identify the best candidate variable at each iteration can be substantially reduced. 

The pseudocode for a lazy greedy variable selection algorithm is given in Algorithm \ref{alg:lazy_greedy}. Here $G^o_U$ is a decreasing-ordered list of the marginal gain bounds for the set of currently unselected candidate variables, $I^o_U$ is the corresponding index set, and $EB_{\text{flag}}^o$ is a Boolean set used to track which entries in the ordered list are exact marginal gains (= 1) and which are upper bounds (= 0). The function $reorder(\cdot)$ is used to denote the insertion of the updated marginal gain value in the appropriate location in the ordered list and the corresponding reordering of the entries in $I^o_U$ and $EB_{\text{flag}}^o$. This step can be efficiently implemented taking advantage of the existing ordering of the lists. 

\begin{algorithm}[!h]
	\caption{Lazy greedy variable selection}
	\begin{algorithmic}[1] 
		\Input $\bm{X}, k$
		\State $I_U=\{1, 2, \dots, v-1, v\};~ I_S = \emptyset$
		\State $G_U = \{g(i)-g(\emptyset), i \in I_U\}$
		\State $[I_U^o, G_U^o]  = sort(G_U)$
		\State $I_S=I_U^o(1);~I_U^o(1)=\emptyset;~ G_U^o(1)=\emptyset$
		\While {$|I_S|<k$}
		\State $EB_{\text{flag}}^o =zeros(size(I_U^o))$
		\While {$EB_{\text{flag}}^o(1) \ne 1$} 
		\State $G_U^o(1) = g(I_S \cup I_U^o(1))-g(I_S)$
		\State $EB^o_{\text{flag}}(1)=1$ 
		\State $[I_U^o, G_U^o, EB_{\text{flag}}^o]  = reorder(G_U^o)$
		\EndWhile
		\State $I_S= I_S \cup I_U^o(1);~ I_U^o(1)=\emptyset;~ G_U^o(1)=\emptyset$
		\EndWhile \\
		\Return $I_S$
	\end{algorithmic}
	\label{alg:lazy_greedy}
\end{algorithm}

\section{\hlight{Candidate Algorithms, Complexity and Performance Metrics}} \label{meth}
\subsection{Candidate Algorithms}
This section provides a description of the six baseline algorithms of our comparative study. Without loss of generality, the data matrix $\bm{X}$ is assumed to have mean-centered columns.

\subsubsection{Forward Selection Component Analysis (FSCA)}
FSCA \citep{IEEEPuggini}, first proposed in \cite{ragnoli2009identifying} in the context of optical emission spectroscopy channel selection, has been found to be an effective tool for data-driven metrology plan optimization in the semiconductor manufacturing industry \citep{mcloone2018methodology,Prakash2012a,susto2019induced}. The algorithm, which can be regarded as the unsupervised counterpart of forward selection regression \citep{bendel1977comparison}, selects, in a greedy fashion, the variables that provide the greatest contribution to the variance in the dataset $\bm{X}$. As depicted in Algorithm \ref{psudocode-FSCA}, FSCA operates recursively selecting, at each iteration, the variable that maximizes the variance explained with respect to the residual matrix $\bm{R}$ obtained by removing from $\bm{X}$ the contribution of the variables selected during the previous iterations (Step 6). The variance explained (VE) performance metric is defined as
\begin{equation}\label{eq:varexp}
V_{\bm{R}} \left(\hat{\bm{R}}(\bm{r}_i) \right) =\Bigg(1 - \frac{||\bm{R} - \hat{\bm{R}}(\bm{r}_i)||_F^2}{||\bm{R}||_F^2}\Bigg) \cdot 100,
\end{equation}
where $\hat{\bm{R}}(\bm{r}_i)$ is the matrix $\bm{R}$ reconstructed by regressing on $\bm{r}_i$, the $i$-th column of $\bm{R}$, i.e.
\begin{equation}
\hat{\bm{R}}(\bm{r}_i) = \frac{\bm{r}_i\bm{r}_i^T}{\bm{r}_i^T \bm{r}_i}\bm{R}.
\label{eq:R}
\end{equation}

\noindent Here, $||\bm{R} - \hat{\bm{R}}(\bm{r}_i)||_F$ is the Frobenius norm of the difference between the reconstructed and actual residual matrix. Maximising the explained variance in (\ref{eq:varexp}) is equivalent to maximising the Rayleigh Quotient of $\bm{RR}^T$ with respect to $\bm{r}_i$ and this can be exploited to achieve a computationally efficient implementation of FSCA as described in \cite{IEEEPuggini}. 

In terms of the greedy search algorithm formulation in Algorithm \ref{alg:general}, FSCA corresponds to setting $g(I_S) =  V_{\bm{X}} (\hat{\bm{X}}(\bm{X}_S))$, where $\bm{X}_S$ is the subset of columns of $\bm{X}$ corresponding to $I_S$ and $\hat{\bm{X}}(\bm{X}_S)$ is the projection of $\bm{X}$ on $\bm{X}_S$, that is:
\begin{equation}
 \hat{\bm{X}}(\bm{X}_S) = \bm{X}_S[\bm{X}_S^T \bm{X}_S]^{-1}\bm{X}_S^T\bm{X}.
 \label{eq:X_reconstruct}
 \end{equation}
A lazy greedy implementation of the FSCA algorithm can then be realised as set out in Algorithm \ref{alg:lazy_greedy}. This will be referred to as L-FSCA, hereafter.

\begin{algorithm}[!h]
\caption{FSCA}
\begin{algorithmic}[1] 
\Input $\bm{X}, k$
\State $I_U=\{1, 2, \dots, v-1, v\};~ I_S = \emptyset$
\State $\bm{R}_1 = \bm{X}$
\For{$j = 1 \text{ to } k$}
	\State $i^{*} = \underset{i \in I_U}{\argmax} \; \Big\{V_{\bm{R}_{j}} (\hat{\bm{R}}_{j}(\bm{r}_i))\Big\}$
	\State $I_S = I_S \cup \{i^*\}; I_U = I_U\setminus \{i^*\}$
	\State $\bm{R}_{j+1} = \bm{R}_{j} - \hat{\bm{R}}_{j}(\bm{r}_{i^*})$
\EndFor \\
\Return $I_S$
\end{algorithmic}
\label{psudocode-FSCA}
\end{algorithm}

\subsubsection{Forward Orthogonal Search Maximizing the Overall Dependency (FOS-MOD)}
FOS-MOD, proposed by \cite{wei2007feature}, selects the most representative variables based on their similarity with the unselected variables using the \emph{squared correlation coefficient}. Given two vectors $\bm{x}_i$ and $\bm{x}_j$, the squared correlation coefficient is defined as
\begin{equation} \label{sqcorr}
\rho^{2}(\bm{x}_i,\bm{x}_j) = \frac{(\bm{x}_i^T\bm{x}_j)^2}{(\bm{x}_i^T\bm{x}_i)(\bm{x}_j^T\bm{x}_j)}.
\end{equation}
The rationale behind the algorithm is to select, at each iteration, the residual direction $\bm{r}_i$ having the highest average $\rho^2$ with the unselected variables. Hence, the variable selection function is defined as
\begin{equation}
  \bar{C}(\bm{r}_i)= \frac{1}{v} \sum_{j=1}^v \rho^{2}(\bm{x}_j, \bm{r}_i).
\label{cost:FOSMOD}
\end{equation}
Note that, since $\bm{r}_i$ is orthogonal to the selected variables, the contribution of these variables to the summation is zero. The pseudocode for FOS-MOD is given in Algorithm \ref{alg:FOS}.

\begin{algorithm}[!h]
\caption{FOS-MOD}
\label{alg:FOS}
\begin{algorithmic}[1] 
\Input $\bm{X}, k$
\State $I_U=\{1, 2, \dots, v-1, v\};~ I_S = \emptyset$
\State $\bm{R}_1 = \bm{X}$
\For{$j = 1 \text{ to } k$}
	\State $i^*=\underset{i \in I_U}{\argmax} \; \{\bar{C}(\bm{r}_i)\}$
	\State $I_S = I_S \cup \{i^*\}; I_U = I_U\setminus \{i^*\}$
	\State $\bm{R}_{j+1} = \bm{R}_j - \hat{\bm{R}}_j(\bm{r}_{i^*})$
\EndFor \\
\Return $I_S$
\end{algorithmic}
\end{algorithm}

\subsubsection{Information Theoretic Feature Selection (ITFS)}
ITFS was originally proposed by \cite{krause2008near} for a sensor placement problem where an initial sensor deployment is exploited to perform a data-driven optimal placement maximising the \emph{mutual information} (MI) between the selected and unselected sensor locations, such that the information lost by removing a sensor at each step is minimized. Specifically, a discrete set of locations $V$ is split into the set of selected $S$ and unselected $U$ locations, respectively. The goal is to place $k$ sensors in order to have the minimal uncertainty over the unselected locations, i.e. 
\begin{equation}
S^* = \underset{|S| = k}{\argmax} ~\{H(U) - H(U | S)\},
\end{equation}
where $H(\cdot)$ is the entropy over a set. Hence, $S^*$ is the set that maximizes the reduction of entropy (uncertainty) over the unselected locations. To achieve a mathematical expression for the MI it is assumed that the variables are distributed according to a multivariate Gaussian distribution. Under this assumption, we can employ a Gaussian process regression model \citep{kersting2007most} to estimate the multivariate distribution of a set of unselected variables $U$ given the selected ones in $S$. For a measurement vector $\bm{s} \in S$ the multivariate distribution of the unmeasured variables is still multivariate Gaussian with mean $\bm{\mu}^*$ and covariance matrix $\bm{\Sigma}^*$, i.e. $\bm{U} \sim \mathcal{N}(\bm{\mu}^*,\bm{\Sigma}^*)$. These are computed as:
\begin{equation} \label{mu}
\bm{\mu}^{*} = \bm{K}(U, S)(\bm{K}(S,S) + \sigma^2 \bm{I})^{-1}\bm{s}
\end{equation}
\noindent and
\begin{equation} \label{sigma}
\bm{\Sigma}^{*} = \bm{K}(U,U) + \sigma^2\bm{I} - \bm{K}(U,S)(\bm{K}(S,S) + \sigma^2 \bm{I})^{-1}\bm{K}(S,U),
\end{equation}
where $\bm{I} \in \mathbb{R}^{|S| \times |S|}$ and $\bm{K}(\cdot , \cdot)$ is a covariance matrix that can be estimated from the available historical data. In general, given two sets of variables $A$ and $B$ indexed by $I_A$ and $I_B$, we have
\begin{equation}
\bm{K}(A,B) = \bm{\Sigma}_{AB} = \frac{1}{m} \bm{X}^{T}_{A} \bm{X}_{B},
\end{equation}
where $\bm{X}_{A} $ and $\bm{X}_{B} $ are the matrices formed by the columns of $\bm{X}$ indexed by $I_A$ and $I_B$, respectively. The hyperparameter $\sigma$ takes into account the measurement noise.
The problem of maximizing the MI belongs to the NP-complete class of problems \citep{krause2008near}, hence an approximated solution can be found using a greedy approach that sequentially selects the variable $\bm{x}_i$ (i.e. the $i$-th column of $\bm{X}$) that maximizes the increment in mutual information \citep{krause2008near}:
\begin{equation} \label{deltami}
\begin{split}
\Delta_{MI}(\bm{x}_i) &= MI( S \cup \bm{x}_i;  U \setminus \bm{x}_i) -  MI( S;  U) \\
				& = H(\bm{x}_i | S) - H(\bm{x}_i | U \setminus \bm{x}_i)
\end{split}
\end{equation}
with the conditional distribution $\bm{x}_i|U \sim \mathcal{N}(\mu^*_U, \sigma^*_U)$, and $\mu^*_U$ and $\sigma^*_U$ estimated via Eq. \ref{mu} and \ref{sigma}. Recalling that for a Gaussian random variable $p$, the entropy $H(p)$ is a function of its variance $\sigma_p$
\begin{equation}
H(p) = \frac{1}{2} ln(2 \pi e \sigma_p),
\end{equation}
Eq. \ref{deltami} can be expressed as
\begin{equation}
\begin{split}
\Delta_{MI}(\bm{x}_i) &= H(\bm{x}_i | S) - H(\bm{x}_i | U \setminus \bm{x}_i)  \\
& = \frac{\sigma_{\bm{x}_i} - \bm{\Sigma}_{S\bm{x}_i}^{T}\bm{\Sigma}_{SS}^{-1}\bm{\Sigma}_{S\bm{x}_i}}{\sigma_{\bm{x}_i} - \bm{\Sigma}_{U\bm{x}_i}^{T}\bm{\Sigma}_{UU}^{-1}\bm{\Sigma}_{U\bm{x}_i}}.
\end{split}
\end{equation} 
The pseudocode for ITFS is given in Algorithm \ref{code:ITFS}. While MI is not in general monotonically increasing, \cite{krause2008near} show that, when $|I_S|<<v$, the diminishing returns property (Eq. \ref{eq:SubmodularityDef}) applies, hence ITFS is guaranteed to be within 63\% of the optimal value in terms of the MI of $\bm{X}$. 


\begin{algorithm}[!h]
\caption{ITFS}
\begin{algorithmic}[1] 
\Input $\bm{X},k$
\State $I_U=\{1, 2, \dots, v-1, v\};~ I_S = \emptyset$
\For{$j = 1 \text{ to } k$}
	\State $i^* = \underset{\bm{x}_i \in U}{\argmax} ~ \Delta_{MI}(\bm{x}_i)$
	\State $I_S = I_S \cup \{i^*\}; I_U = I_U\setminus \{i^*\}$
\EndFor \\
\Return $I_S$
\end{algorithmic}
\label{code:ITFS}
\end{algorithm}

\subsubsection{Principal Feature Selection (PFS)}
PFS, presented in  Algorithm \ref{code:OPFS},  is a PCA guided approach to variable selection introduced by \cite{cui2008orthogonal}. At each iteration it selects the variable that corresponds to the residual vector $\bm{r}_i$ that is most correlated with the first principal component (PC) $\bm{p}_1$ of the residual matrix $\bm{R}$ obtained after the contribution of the variables selected in the previous iterations has been removed (as per the matrix deflation Step at line 7 with $\hat{\bm{R}}$ defined as in Eq. \ref{eq:R}). The correlation is measured using Pearson's correlation coefficient which, for mean-centred residual vector $\bm{r}_i$ and principal component $\bm{p}_1$, is defined as
\begin{equation}
\rho(\bm{r}_i,\bm{p}_1) = \frac{\bm{r}_i^T\bm{p}_1}{\sqrt{(\bm{r}_i^T\bm{r}_i)(\bm{p}_1^T\bm{p}_1)}}
\end{equation}

\begin{algorithm}[!h]
\caption{PFS}
\begin{algorithmic}[1] 
\Input $\bm{X},k$
\State $I_U=\{1, 2, \dots, v-1, v\};~ I_S = \emptyset$
\State $\bm{R}_1 = \bm{X}$
\For{$j = 1 \text{ to } k$}
	\State $\bm{p}_1 =$ First principal component of $\bm{R}_j$
	\State $i^{*} = \underset{i \in I_U}{\argmax} ~ |\rho(\bm{r}_i,	\bm{p}_1)|$
	\State $I_S = I_S \cup \{i^*\}; I_U = I_U\setminus \{i^*\}$
	\State $\bm{R}_{j+1} = \bm{R}_j - \hat{\bm{R}}_j(\bm{r}_{i^*})$
\EndFor \\
\Return $I_S$
\end{algorithmic}
\label{code:OPFS}
\end{algorithm}

\subsubsection{Forward Selection Frame Potential (FSFP)}
The FSFP algorithm introduced in \cite{zocco2017mean} was inspired by the frame potential based sensor selection method proposed by \cite{ranieri2014near} for linear inverse problems.  The \emph{frame potential} (FP) of a matrix $\bm{X}$ is defined as
\begin{equation}
\label{FP-def}
FP(\bm{X}) = \sum_{i,j =1}^{v} | \left\langle \bm{x}_i, \bm{x}_j \right\rangle|^2
\end{equation}
where $\bm{x}_j$ denotes the $j$-th column of $\bm{X}$. It is an attractive metric for variable selection as minimising FP encourages orthogonality among the selected columns of $\bm{X}$. \cite{ranieri2014near} showed that the set function $f(I_U)=FP(\bm{X}) - FP(\bm{X}_{S})$, where $I_S =I_X \setminus I_U$, is monotone increasing, submodular and $f(\emptyset) = 0$. 
Therefore it satisfies the conditions for Nemhauser's bound (Eq. \ref{eq:SubModularOptBound}) and its greedy maximisation is guaranteed to be within 63\% of the global optimum solution. However, with this formulation variable selection is achieved by backward elimination, that is, the algorithm begins with all variables selected and then unselects variables one at a time until $k$ variables remain. Therefore, the algorithm must iterate $v-k$ times, at which point the desired variables are given by $I_S=I_{X} \setminus I_U$.  Backward elimination is computationally much more demanding than forward selection, especially when $k<<v$.  In our previous work \citep{zocco2017mean} we propose an FP based forward selection algorithm (i.e. FSFP) where the variable selection metric is given by 
\begin{equation}
g(I_S)=FP(\bm{X}) - FP(\bm{X}_{S}). 
\end{equation}
The resulting algorithm (summarised in Algorithm \ref{code-FSFP}) is computationally much more efficient than backward elimination, but no longer meets the requirements for the $\mathcal{B}_{\text{N}}$ bound (Eq. \ref{eq:SubModularOptBound}) since $g(\cdot)$ is a monotone decreasing function and $g(\emptyset) \ne 0$.  However, submodularity does apply allowing a lazy greedy implementation. 

As suggested in \cite{ranieri2014near}, FP-based greedy selection provides better results when matrix $\bm{X}$ is normalized to have columns with unitary norm. However, this presents a challenge for FSFP as all variables have the same FP, precisely equal to one, leading to an ambiguous choice for the first variable. Different solutions can be adopted to address this ambiguity. The simplest solution is to randomly select the first variable, however this leads to an output $I_S$ that is not unique for a given $\bm{X}$. Two alternatives are to select the optimum pair of variables instead of a single variable at the first step, or to evaluate FSFP considering all the possible variables as first choice and then choosing the best one, but both these methods lead to a huge increase in computational cost. The approach we have chosen instead is to select the first variable using FSCA. The resulting algorithm, denoted as FSFP-FSCA, is given in Algorithm \ref{FSFP-FSCA}.  

\begin{algorithm}[!h]
\caption{FSFP}
\begin{algorithmic}[1] 
\Input $\bm{X},k$
\State $I_U=\{1, 2, \dots, v-1, v\};~ I_S = \emptyset$
\For{$j = 1 \text{ to } k$}
	\State $i^* = \underset{i \in I_U}{\argmax} \hspace{1pt} \{FP(\bm{X}) - FP(\bm{X}_{S \cup \left\lbrace \bm{x}_i \right\rbrace})\}$
	\State $I_S = I_S \cup \{i^*\}; I_U = I_U\setminus \{i^*\}$
\EndFor \\
\Return $I_S$
\end{algorithmic}
\label{code-FSFP}
\end{algorithm}

\begin{algorithm}[!h]
\caption{FSFP-FSCA}
\begin{algorithmic}[1] 
\Input $\bm{X},k$
\State Normalize $\bm{X}$ to have unitary norm columns
\State $I_S = FSCA(\bm{X}, 1); I_U =\{1, 2, \dots, v-1, v\}\setminus I_S$
\For{$j = 2 \text{ to } k$}
	\State $i^* = \underset{i \in I_U}{\argmax} \hspace{1pt} \{FP(\bm{X}) - FP(\bm{X}_{S \cup \left\lbrace \bm{x}_i \right\rbrace})\}$
	\State $I_S = I_S \cup \{i^*\}; I_U = I_U\setminus \{i^*\}$
\EndFor \\
\Return $I_S$
\end{algorithmic}
\label{FSFP-FSCA}
\end{algorithm}

\subsubsection{Unsupervised Forward Selection (UFS)}
\cite{whitley2000unsupervised} proposed a feature selection method in the context of computational chemistry and drug design, which they refer to as unsupervised forward selection (UFS). The effect of an agent on a biological system can be evaluated through bioassays to capture the responses of process variables to stimuli. The large number of variables involved often makes the identification of relationships among them challenging, therefore UFS has been developed to eliminate feature redundancy by identifying a reduced set of relevant system properties. The selection metric employed is the \emph{squared multiple correlation coefficient} with respect to an orthonormal basis spanning the columns of $\bm{X}$ corresponding to the already selected variables, $\bm{X}_{S}$, with the orthonormal basis computed by, for example, the Gram-Schmidt procedure. Denoting the orthonormal basis corresponding to $\bm{X}_{S}$ as $\bm{C}_S=[\bm{c}_1, \bm{c}_2, \dots, \bm{c}_{|I_S|}]$, the squared multiple correlation coefficient $R^2$  of vector $\bm{x}_i$, $i \in I_U$, is defined as
\begin{equation}
	R^2(\bold{x}_i, \bold{C}_S) =|\sum_{j=1}^{|I_S|}{\bold{c}_j^T \bold{x}_i  \bold{c}_j}|^2  = \sum_{j=1}^{|I_S|}(\bold{c}_j^T \bold{x}_i)^2 = \bold{x}_i^T\bold{C}_S\bold{C}_S^T \bold{x}_i
\end{equation}
Using this metric the UFS algorithm selects at each iteration the variable with the smallest $R^2$, while discarding variables whose $R^2$ value exceeds a user defined similarity threshold $R^2_{max}$, until all variables have either been selected or discarded. To maintain consistency with the forward greedy variable selection algorithm structure considered in this paper (Algorithm \ref{alg:general}), in our implementation of UFS we omit the similar variable pruning step and instead terminate when $k$ variables have been selected, as described in Algorithm \ref{UFS-code}.

\begin{algorithm}[!h]
\caption{UFS}
\label{UFS-code}
\begin{algorithmic}[1] 
\Input $\bm{X},k$
\State Normalize $\bm{X}$ to have unitary norm columns
\State $\bm{Q} = \bm{X}^T \bm{X}$
\State $I_S = \{i_1, i_2\}$, where $i_1$ and $i_2$ ($i_1 < i_2$) are the row and column indices of the element of $\bm{Q}$ with the smallest absolute value
\State Define an orthonormal basis of the variables indexed by $I_S: \bm{C}_{S_2}=[\bm{c}_1, \bm{c}_2]$ where  $\bm{C}_{S_2} =$ \emph{Gram-Schmidt}($\bm{X}_{S}$)
\For{$j = 3 \text{ to } k$}
    \State $i^* = \underset{i \in I_U}{\argmin} ~ R^2(\bm{x}_i, \bm{C}_{S_{j-1}})$
	\State $I_S = I_S \cup \{i^*\}; I_U = I_U\setminus \{i^*\}$
	 \State $\bm{C}_{S_j} =$ \emph{Gram-Schmidt}($\bm{X}_{S}$) 
\EndFor \\
\Return $I_S$
\end{algorithmic}
\end{algorithm}

The UFS algorithm is equivalent to defining the selection metric $g(\cdot)$ in Algorithm \ref{alg:general} as 
\begin{equation}
	g(I_S) = - \sum_{j=1}^{|I_S|} \sum_{i=1}^{j-1} (\bold{c}_i^T \bold{x}_j)^2
	\label{eq:UFS-set_function}
\end{equation}
which corresponds to defining the marginal gain as $-R^2(\bold{x}, \bold{C}_{S_k})$ and replacing $argmin$ with $argmax$ at step 6 of Algorithm \ref{UFS-code}. It can be shown that this metric is a submodular function, hence an exact lazy greedy implementation of UFS is possible. However, $g(I_S)$ is a monotone decreasing function, rather than increasing, and therefore the $\mathcal{B}_{\text{N}}$ performance bound (Eq. \ref{eq:SubModularOptBound}) does not apply.

\subsection{Computational Complexity}
\label{sec:compComp}

All the algorithms considered share a similar greedy selection structure, and all rely on the existence of linear dependency between the selected and unselected variables.  The element that distinguishes one algorithm from another is the variable selection criterion employed (see Table \ref{tab:selection_criteria}).   The differences in these criteria result in algorithms with substantially different computational complexity, as summarised in Table \ref{TableComplexity}. 
%
%
FSCA and FOS-MOD have the same order of magnitude of computational complexity with the only difference between them being an additional dot product term in the FOS-MOD metric, as discussed in \cite{IEEEPuggini}. Whenever computing FSCA or FOS-MOD is prohibitive, PFS can be employed in conjunction with the Nonlinear Iterative Partial Least Squares (NIPALS) algorithm \citep{wold1973nonlinear} to compute the first principal component, yielding a $O((4N+8)kmv) \rightarrow O(Nkmv)$ complexity algorithm \citep{IEEEPuggini}. Here $N$ denotes the average number of iterations for the NIPALS algorithm to converge to the first principle component. The lazy greedy formulation of FSCA achieves a similar complexity to PFS.   

The term $O(mv^2)$ arises in FSFP-FSCA due to the complexity of selecting the first variable using FSCA, whereas in both ITFS and UFS it is associated with the initial computation of the $v \times v$ covariance matrices, $\bm{\Sigma}_{UU}$ and $\bm{Q}$, respectively.  ITFS is dominated by the $O(v^3)$ inverse covariance calculation which is repeated approximately $v$ times per iteration and for each of the $k$ variables selected, hence the $O(kv^4)$ complexity of this algorithm. A lazy greedy implementation reduces this to $O(kv^3)$ \citep{krause2008near}. In the FSFP-FSCA and UFS algorithms the FP and  $R^2$ variable selection metrics have  $O(j^2mv)$ and $O(jmv)$ complexity, respectively, at the $j$-th iteration, hence the accumulated complexity for these operations over $k$ iterations is $O(k^3mv)$ and $O(k^2mv)$, respectively. Since FP is submodular, a lazy implementation of FSFP-FSCA has complexity $O(mv^2 + k^3m)$ as the first selection performed via FSCA requires $O(mv^2)$ and the subsequent iterations require $O(k^3m)$. Similarly, UFS has an $O(mv^2 + k^2m)$ lazy implementation. The formulation of the PFS algorithm is not compatible with a lazy implementation. Also, a greedy implementation of the non-submodular FOS-MOD algorithm has not been reported in the literature and its feasibly  has not been investigated in this work, hence it is recorded as ``not investigated'' in the table.


\begin{table}
	\centering
	\caption{The selection criterion corresponding to each considered algorithm.\\[-0.1in]}
	\label{tab:selection_criteria}
	\begin{tabular}{l@{\hskip 0.1in}l}
		\hline\noalign{\smallskip}
		Method name & Selection criterion for $i \in I_U \rightarrow I_S$ \\ 
		\noalign{\smallskip}
		\hline
		\noalign{\smallskip}
		{FSCA}  & Max. explained variance with respect to $\bm{X}$ (or $\bm{R}$)\\ 
		{L-FSCA}  & Max. explained variance with respect to $\bm{X}$ (or $\bm{R}$) \\ 
		{PFS} &  Max. correlation with the first PC of the residual matrix $\bm{R}$\\ 
		{FOS-MOD} &  Max. squared correlation with respect to $\bm{X}$ (or $\bm{R}$)\\ 
		{FSFP-FSCA} & Min. frame potential of the selected variables $\bm{X}_{S}$ \\ 
		{ITFS} &  Max. mutual information between $\bm{X}_{S}$ and $\bm{X}_{U}$\\ 
		{UFS}  & Min. correlation with the selected variables $\bm{X}_{S}$ \\ 
		\hline
	\end{tabular}
\end{table}

\begin{table}
\centering
\caption{Computational complexity of the variable selection methods under consideration (when selecting $k$ variables): $k$ is the number of selected variables, $v$ is the number of candidate variables, $m$ is the number of measurements (observations) and $N$ is the mean number of iterations for the PCA algorithm to converge to the first principal component in PFS.\\[-0.1in]} 
\label{TableComplexity}
\begin{tabular}{lcc}
\hline\noalign{\smallskip}
 Algorithm  & Complexity ($k\ll v$)  & Lazy Implementation  \\
\noalign{\smallskip}
\hline
\noalign{\smallskip}
FSCA & $O(kmv^{2})$  & $\rightarrow O(kmv)$   \\
FOS-MOD & $O(kmv^{2})$  & not investigated  \\
PFS (NIPALS) & $O(Nkmv)$ & not applicable \\
ITFS & $O(mv^2 + kv^4)$   &  $\rightarrow O(mv^2 + kv^3)$  \\
FSFP-FSCA &  $O\big(mv^{2}+k^3mv\big)$   & $\rightarrow O(mv^2 + k^3m)$ \\   
UFS & $O\big(mv^2 + k^2mv)$  & $\rightarrow O\big(mv^2 + k^2m)$ \\
\hline
\end{tabular}
\end{table}

\subsection{\hlight{Performance Metrics}}
The comparison is based on six metrics. Three of these are the variance explained $V_{\bm{X}}(\bm{X}_S)$, the frame potential $FP(\bm{X}_S)$ and the mutual information $MI(\bm{X}_S)$, as previously defined in Eq. (\ref{eq:varexp}), (\ref{FP-def}) and  (\ref{deltami}), respectively. These are complemented by three metrics introduced to characterise the evolution of $V_{\bm{X}}(\bm{X}_{S})$ with the number of selected variables $k$, namely, the area under the variance curve $AUC$, the relative performance $r$, and the number of variables needed to reach $n\%$ variance explained $k_{n\%}$. $AUC$ is defined as 
\begin{equation}
AUC = \frac{0.01}{v-1}\sum_{k=1}^{v-1} V_{\bm{X}}(\hat{\bm{X}}_{k})
\end{equation}
where $\hat{\bm{X}}_{k}$ denotes the approximation of $\bm{X}$ with  $\bm{X}_S$, where $|I_S|=k$, as defined in Eq. (\ref{eq:X_reconstruct}). The factor $100(v-1)$ normalises the expression to yield a maximum value of 1 when 100\% of the variance is explained by a single variable. The relative performance $r$ expresses how well an algorithm performs relative to the other algorithms over the first $k^*_{99\%}$ selected variables. Specifically, it is the percentage of times over the interval $k=1, ..., k^*_{99\%}$  that the algorithm is the top ranked, i.e. yields the maximum value of $V_{\bm{X}}(\hat{\bm{X}}_{k})$, that is,
\begin{equation}
r = \frac{100}{k^*_{99\%}} \sum_{k=1}^{k^*_{99\%}} is\_top\_rank(k).
\end{equation}
Here $k^*_{99\%}$ denotes the minimum value of $k$ required to have all the algorithms exceeding $99\%$ variance explained and $is\_top\_rank(k)$ is 1 if the algorithm is the best among those being evaluated and zero otherwise, for $|I_S|=k$. Finally, $k_{n\%}$ provides a measure of the level of compression provided for a given target reconstruction accuracy $n$, and is defined as 
\begin{equation}
k_{n\%}=\underset{k}{\argmin}{~V_{\bm{X}}(\hat{\bm{X}}_{k})} ~~ s.t. ~~V_{\bm{X}}(\hat{\bm{X}}_{k}) \ge n.
\end{equation}
Note that the $AUC$, $r$ and $k_{n\%}$ metrics are functions of VE, hence algorithms optimizing VE are implicitly optimizing them too. The choice of having four out of six metrics based on VE is motivated by the fact that, in contrast to MI and FP, VE is a direct measurement of the approximation error $\bm{X} - \hat{\bm{X}}$ which, in practice, is the quantity we wish to minimize.

\section{\hlight{Results}} \label{res}
This section compares the performance of the algorithms described above, namely FSCA, L-FSCA, FOS-MOD, PFS, ITFS, FSFP-FSCA and UFS, on the simulated and real world datasets summarized in Table \ref{tab:datasetsSummary}. For reasons of space, in this section FOS-MOD will be referred to as FOS, FSFP-FSCA as FP-CA and L-FSCA as L-CA. The code that generated the results is written in MATLAB and is publicly available\footnote{\url{https://github.com/fedezocco/GreedyVarSel-MATLAB}} to facilitate further benchmarking of greedy search algorithms for unsupervised variable selection. Our experiments were conducted using Matlab R2019b running on an Intel(R) Core(TM) i7-9700K CPU @ 3.60GHz  with 32 GB  RAM.

\begin{table*}
\begin{center}
\caption{Overview of the case study datasets.\\[-0.1in]}
\label{tab:datasetsSummary}
\begin{threeparttable}
\begin{tabular}{l@{\hskip 0.6in}l@{\hskip 0.6in}c@{\hskip 0.6in}c} 
\hline
Size & Dataset & $m$ & $v$ \\
\hline
\multirow{8}*{Smaller} & Simulated 1 \citep{IEEEPuggini} & 1000 & 26 \\ 
& Simulated 2 \citep{IEEEPuggini} & 1000 & 50 \\ 
& Pitprops \citep{Jeffers} & 180 & 13 \\ 
& Semiconductor \citep{Prakash2012a} & 316 & 50 \\ 
& Arrhythmia\tnote{1} ~\citep{guvenir1997supervised} & 452 & 258 \\ 
& Sales\tnote{2} ~\citep{tan2014time} & 812 & 52 \\
& Gases\tnote{3} ~\citep{vergara2012chemical} & 1586 & 129 \\
& Music\tnote{4} ~\citep{zhou2014predicting} & 1059 & 68 \\
\hline
\multirow{4}*{Larger} & \hlight{PlasmaEtch \citep{puggini2018_anomaly}} & 2194 & 2046\\
& \hlight{YaleB\tnote{6} ~\citep{georghiades2001_YaleB}} & 2414 & 1024\\
& \hlight{WaferIPCM\tnote{5} ~\citep{olszewski2001generalized}} & 7164 & 152\\
& \hlight{USPS\tnote{6} ~\citep{hull1994USPS}} & 9298 & 256\\
\hline
\end{tabular}
\begin{tablenotes}\footnotesize
\item[1] \url{https://archive.ics.uci.edu/ml/datasets/Arrhythmia}
\item[2] \url{https://archive.ics.uci.edu/ml/datasets/Sales_Transactions_Dataset_Weekly}
\item[3] \url{https://archive.ics.uci.edu/ml/datasets/Gas+Sensor+Array+Drift+Dataset+at+Different+Concentrations}
\item[4] \url{https://archive.ics.uci.edu/ml/datasets/Geographical+Original+of+Music}
\item[5] \url{http://timeseriesclassification.com/description.php?Dataset=Wafer}
\item[6] \url{http://www.cad.zju.edu.cn/home/dengcai/Data/data.html}
\end{tablenotes}
\end{threeparttable}
\end{center}
\end{table*}

\subsection{Preliminary Overview}
Table \ref{tab:summary} gives a preliminary overview of the performance of the algorithms on each \hlight{of the smaller datasets}. The metrics reported are $AUC$ and $k_{n\%}$ with $n \in$ \{95, 99\}. $k_{n\%}$ suggests that FSCA and its lazy version provide the highest level of compression in all datasets with the exception of $k_{95\%}$ in `Music' and `Simulated 1'. The lazy version of FSCA achieves the same performance as FSCA on all eight datasets, confirming that, while VE is not submodular, it essentially behaves in a submodular fashion with respect to greedy selection. An inspection of the sequence of variables selected by each algorithm showed that small deviations did occur with L-FSCA for the `Sales', `Arrhythmia' and `Pitprops' datasets. It produced a different permutation in the sequence of selected variables between $k=9$ and $k=30$ with `Sales' and between $k=25$ and $k=29$ with  `Arrhythmia'. This resulted in a maximum VE deviation of 0.006 within these intervals, but no deviation thereafter. In the `Pitprops' dataset L-FSCA selected a different variable at $k=10$ resulting in a reduction in VE of 0.006 at that point. However, the net impact when the 11th variable is included is a positive deviation (increase in VE) of 0.04. FSCA methods show 13 values of $k_{n\%}$ in bold, followed by PFS with 12, ITFS with 8, FOS with 6, FP-CA with 3 and UFS with 0. The area under the VE curve, i.e. $AUC$, is greatest with FSCA methods for 6 datasets, followed by PFS with 5, FOS with 2 and 0 for the remaining algorithms. `Pitprops' and `Music' are the datasets most difficult to compress to a lower dimension as they provide the lowest values of $AUC$ (0.756 and 0.785, respectively). In contrast, despite having the second highest dimension $v$, `Gases' can be represented by just 3 variables with 99\% accuracy.
\begin{table*}
\centering
\caption{A summary of the performance of the different selection algorithms \hlight{ on the smaller datasets}. Best values are in bold.\\[-0.1in]}
\label{tab:summary}
\begin{tabularx}{\textwidth}{c@{\hskip 0.11in}c@{\hskip 0.11in}c@{\hskip 0.11in}c@{\hskip 0.11in}c@{\hskip 0.11in}c@{\hskip 0.11in}c@{\hskip 0.11in}c@{\hskip 0.11in}c}\hline
 Dataset             & Metric & FSCA & L-CA & FP-CA & PFS & ITFS  & FOS & UFS \\ \hline
\multirow{3}*{\makecell{Semic. \\ \scriptsize(316 $\times$ 50)}} & $k_{95\%}$ & \textbf{4}    & \textbf{4}    & 6    & \textbf{4}    & \textbf{4}  & \textbf{4}    & 5                         \\
              & $k_{99\%}$ & \textbf{7}    & \textbf{7}    & 10   & \textbf{7}    & \textbf{7}  & \textbf{7}    & 9                                       \\
              & $AUC$    & \textbf{0.976} & \textbf{0.976} & 0.969 & \textbf{0.976} & 0.970 & 0.975 & 0.961                           \\ \hline
\multirow{3}*{\makecell{Pitpr. \\ \scriptsize(180 $\times$ 13)}}  & $k_{95\%}$    & \textbf{9}    & \textbf{9}    & \textbf{9}    & \textbf{9}    & \textbf{9}  & \textbf{9}    & 10                           \\
              & $k_{99\%}$ & 11   & 11   & 11   & 11   & 11 & 11   & 11                           \\
              & $AUC$    & \textbf{0.756} & \textbf{0.756} & 0.726 & \textbf{0.756} & 0.730 & \textbf{0.756} & 0.683                           \\ \hline
\multirow{3}*{\makecell{Arryt. \\ \scriptsize(452 $\times$ 258)}}     & $k_{95\%}$ & \textbf{47}   & \textbf{47}   & 241  & \textbf{47}   & 144  & 178  & 193                         \\
              & $k_{99\%}$ & \textbf{71}   & \textbf{71}   & 256  & \textbf{71}   & 173  & 240  & 239                                      \\ 
              & $AUC$    & \textbf{0.954} & \textbf{0.954} & 0.688 & 0.952 & 0.844 & 0.849 & 0.741      \\ \hline
\multirow{3}*{\makecell{Sales \\ \scriptsize(812 $\times$ 52)}}         & $k_{95\%}$ & \textbf{5}    & \textbf{5}    & 7    & \textbf{5}    & \textbf{5}    & 9  & 13                             \\ 
              & $k_{99\%}$ & \textbf{38}   & \textbf{38}   & 41   & 39   & 40   & 42 & 42                            \\
              & $AUC$    & \textbf{0.976} & \textbf{0.976} & 0.975 & \textbf{0.976} & 0.975 & 0.970 & 0.968                           \\ \hline
\multirow{3}*{\makecell{Gases \\ \scriptsize(1586 $\times$ 129)}}         & $k_{95\%}$ & \textbf{2}    & \textbf{2}    & \textbf{2}    & \textbf{2}    & \textbf{2}    & 3    & 4                          \\
              & $k_{99\%}$ & \textbf{3}    & \textbf{3}    & 6    & \textbf{3}    & 11   & 9    & 17                            \\
              & $AUC$    & \textbf{0.999} & \textbf{0.999} & 0.998 & \textbf{0.999} & 0.992 & 0.991 & 0.988                            \\ \hline
\multirow{3}*{\makecell{Music \\ \scriptsize(1059 $\times$ 68)}}         & $k_{95\%}$ & 49   & 49   & 53   & \textbf{48}   & 49   & 49   & 56                                   \\
              & $k_{99\%}$ & \textbf{61}   & \textbf{61}   & 64   & \textbf{61}   & \textbf{61}   & \textbf{61}   & 64                                       \\
              & $AUC$    & 0.785 & 0.785 & 0.758 & \textbf{0.786} & 0.771 & 0.785 & 0.724                               \\ \hline
\multirow{3}*{\makecell{Sim. 1 \\ \scriptsize(1000 $\times$ 26)}}   & $k_{95\%}$ & 5    & 5    & \textbf{4}    & 5    & \textbf{4}    & \textbf{4}    & 15                           \\
              & $k_{99\%}$ & \textbf{6}    & \textbf{6}    & 7    & \textbf{6}    & \textbf{6}    & \textbf{6}    & 21                                       \\
              & $AUC$    & 0.935 & 0.935 & 0.935 & 0.935 & 0.934 & \textbf{0.937} & 0.779                            \\ \hline
\multirow{3}*{\makecell{Sim. 2 \\ \scriptsize(1000 $\times$ 50)}}   & $k_{95\%}$ & \textbf{18}   & \textbf{18}   & 24 & 19   & 23   & 19   & 25                              \\
              & $k_{99\%}$ & \textbf{22}   & \textbf{22}   & 25 & \textbf{22}   & 25   & 23   & 25                            \\
              & $AUC$    & \textbf{0.863} & \textbf{0.863} & 0.770 & 0.855 & 0.812 & 0.852 & 0.750		 \\ \hline                          
\end{tabularx}
\end{table*}

The indices of the variables selected by each algorithm for selected datasets are shown in Table \ref{SelectionOrderSelectedDataSets}. Each table entry is the index of the variable selected at the $k$-th iteration. The sequence of variables selected by a given algorithm is therefore defined by the corresponding table column.  In all cases FSCA and its lazy implementation L-CA select the same variables which further sustains the hypothesis that variance explained is behaving as a submodular function with these datasets. The frame-potential-based algorithm, i.e. FP-CA, selects the same first variable as FSCA because, as discussed in the previous section, FSCA is used for $k = 1$ before switching to a FP-based selection when $k > 1$. The FP-based selection overall does not show similarities with MI-based and VE-based selections. In `Simulated 1' PFS, FSCA, L-CA, ITFS and FOS select the same set when $k$ = 6, but the variables are selected with a different order. In `Sales', five out of the six variables selected by ITFS are in common with FSCA/L-CA. Moreover, for $k = 6$, UFS has five variables in common with FOS and FP-CA. In `Simulated 2' and `Semiconductor': the only relevant similarity among the algorithm selections is in the latter, where FSCA/L-CA and PFS share three variables when $k = 5$. 
\begin{table*}
\centering
\caption{Ordered indices of the selected variables for four datasets when $k$ = 1, ..., 6. For each value of $k$, the $k-$th selected variable is indicated.\\[-0.1in]}
\label{SelectionOrderSelectedDataSets}
\begin{tabularx}{\textwidth}{c@{\hskip 0.22in}c@{\hskip 0.22in}c@{\hskip 0.22in}c@{\hskip 0.22in}c@{\hskip 0.22in}c@{\hskip 0.22in}c@{\hskip 0.22in}c@{\hskip 0.22in}c}
\hline
Dataset & $k$ & FSCA & L-CA & FP-CA & PFS & ITFS & FOS & UFS\\
\hline
\multirow{6}*{Sim. 1}  & 1 & \{26\} & \{26\} & \{26\} & \{25\} & \{1\} & \{25\} & \{26\}\\ 
& 2 & \{25\} & \{25\} & \{1\} & \{26\} & \{13\} & \{13\} & \{1\}\\
& 3 & \{19\} & \{19\} & \{9\} & \{19\} & \{7\} & \{19\} & \{5\}\\
                          & 4   & \{1\}  & \{1\}  & \{22\} & \{1\}  & \{19\} & \{1\}  & \{2\} \\ 
                          & 5   & \{7\}  & \{7\}  & \{17\} & \{7\}  & \{25\} & \{7\}  & \{6\} \\
                          & 6   & \{13\} & \{13\} & \{25\} & \{13\} & \{26\} & \{26\} & \{4\} \\     
\hline
\multirow{6}*{Sim. 2}  & 1   & \{49\}         & \{49\}         & \{49\}       & \{38\}     & \{49\} & \{34\} & \{22\}\\ 
& 2 & \{35\} & \{35\} & \{25\} & \{35\} & \{19\} & \{28\} & \{6\}\\
& 3 & \{29\} & \{29\} & \{10\} & \{37\} & \{29\} & \{31\} & \{11\}\\
                          & 4  & \{31\} & \{31\} & \{18\} & \{49\} & \{12\} & \{43\} & \{21\} \\
                          & 5  & \{45\} & \{45\} & \{8\}  & \{36\} & \{1\}  & \{26\} & \{24\}  \\
                          & 6  & \{27\} & \{27\} & \{19\} & \{39\} & \{16\} & \{47\} & \{7\} \\         
\hline
\multirow{6}*{Semic.}  & 1   & \{45\}         & \{45\}         & \{45\}       & \{45\}     & \{35\} & \{42\} & \{14\}\\ 
& 2 & \{27\} & \{27\} & \{14\} & \{27\} & \{15\} & \{27\} & \{12\}\\
& 3 & \{1\} & \{1\} & \{11\} & \{23\} & \{25\} & \{1\} & \{5\}\\
                          & 4   & \{24\} & \{24\} & \{4\} & \{21\} & \{46\} & \{21\} & \{24\} \\ 
                          & 5   & \{9\}  & \{9\}  & \{1\} & \{9\}  & \{31\} & \{38\} & \{49\} \\
                          & 6   & \{49\} & \{49\} & \{9\} & \{16\} & \{16\} & \{46\} & \{26\} \\      
\hline
\multirow{6}*{Sales}  & 1   &  \{48\}        & \{48\}         & \{48\}       & \{48\}     & \{48\} & \{42\} & \{48\}\\
& 2 & \{38\} & \{38\} & \{1\} & \{38\} & \{9\} & \{1\} & \{1\}\\
& 3 & \{9\} & \{9\} & \{2\} & \{17\} & \{49\} & \{4\} & \{2\}\\ 
                          & 4   &  \{52\} & \{52\} & \{52\} & \{51\} & \{52\} & \{2\} & \{3\} \\
                          & 5   & \{17\} & \{17\} & \{3\}  & \{9\}  & \{17\} & \{3\} & \{4\} \\ 
                          & 6     &   \{49\} & \{49\} & \{4\}  & \{43\} & \{45\} & \{5\} & \{5\} \\   
\hline

\end{tabularx}
\end{table*}

\subsection{Simulated Datasets}
\textbf{Simulated dataset 1.} In this dataset, which is from \cite{IEEEPuggini}, a set of 4 i.i.d. variables $w,x,y,z \sim \mathcal{N}(0, 1)$ and 22 noise variables,  $\epsilon_i \sim \mathcal{N}(0, 0.1)$, $i=1,...,20$ and $\epsilon_{21}$, $\epsilon_{22} \sim \mathcal{N}(0, 0.4)$ are used to create a set of dependent variables $w_i = w + \epsilon_i$, $x_i = x + \epsilon_{i+5}$, $y_i = y + \epsilon_{i+10}$, $z_i = z + \epsilon_{i+15}$, for $i= 1, \dots, 5$, and  $h_1 = w + x + \epsilon_{21}$, $h_2 = y + z + \epsilon_{22}$. The final matrix is then defined as
\begin{equation}
 \bm{X} = [\bm{w}, \bm{w}_1, \dots, \bm{w}_5,~\bm{x}, \bm{x}_1, \dots, \bm{x}_5,~ \bm{y}, \bm{y}_1, \dots, \bm{y}_5,~ \bm{z}, \bm{z}_1, \allowbreak \dots, \bm{z}_5,~ \bm{h}_1, \bm{h}_2]
\end{equation}
with the columns generated as $m=1000$ realizations of the random variables, yielding a dataset $\bm{X} \in \mathbb{R}^{1000 \times 26}$. 

\begin{figure}
\centering
\includegraphics[width=0.8\textwidth]{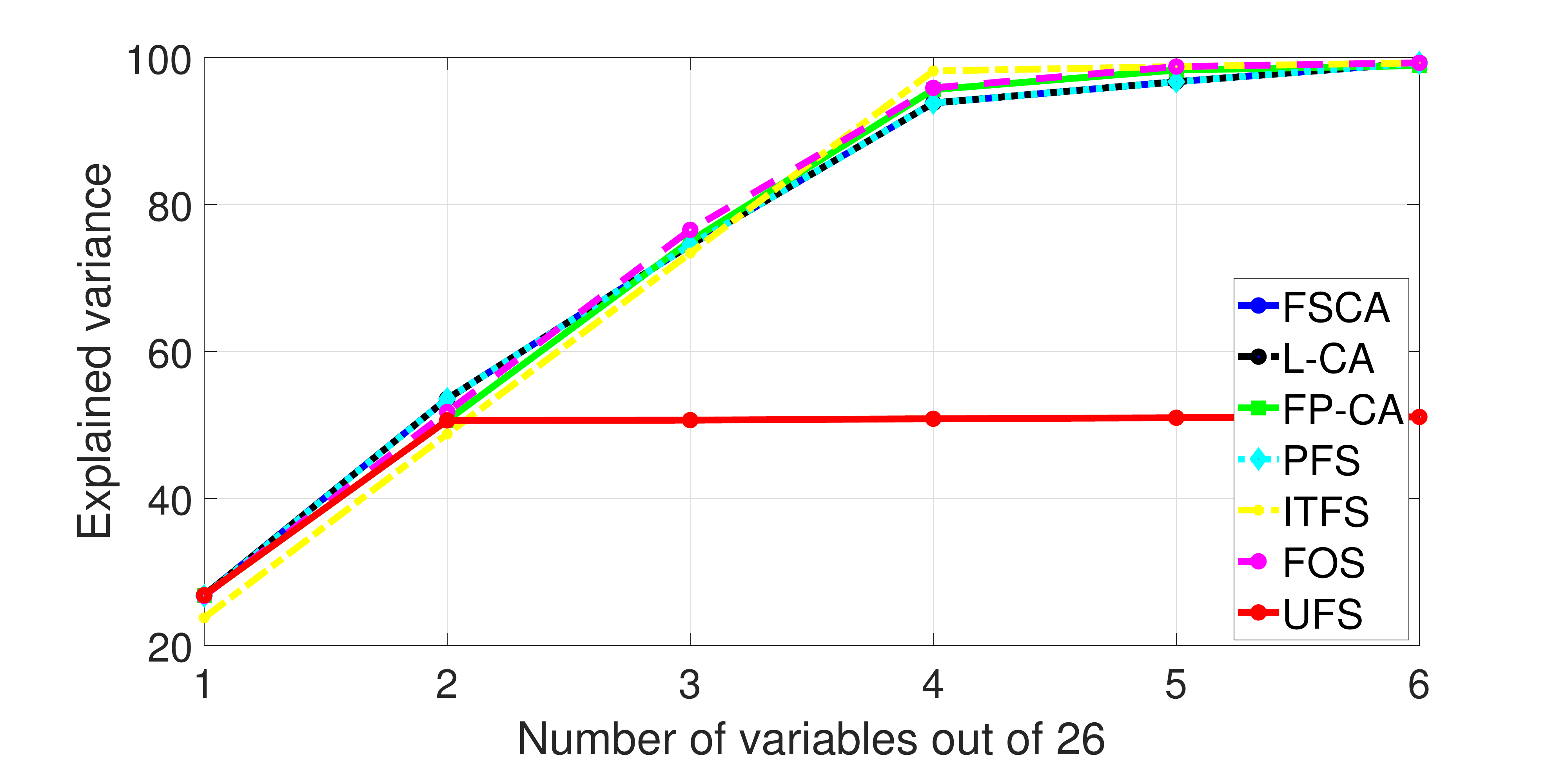} 
\caption{Variance explained as a function of the number of selected variables, $k$, for the `Simulated 1' case study}
\label{fig:sim1}
\end{figure}

Figure \ref{fig:sim1} shows the explained variance as a function of the number of selected variables for each algorithm. With the exception of UFS, all the methods achieve a VE greater than 90\% with 4 variables. Recalling Table \ref{tab:summary}, in this case FOS is the best according to all three metrics, whereas UFS requires more than 6 variables to exceed 95\% VE, which is beyond the figure horizontal axis limit. As observed in Table \ref{SelectionOrderSelectedDataSets}, in this dataset the algorithms select similar variable sequences with the exception of UFS.

\vspace{2mm}
\noindent \textbf{Simulated dataset 2.} This dataset is also taken from \cite{IEEEPuggini} and is characterized by having two blocks of variables. The first block $\bm{X}^I$ is composed of $u$ independent variables, while the second block $\bm{X}^D$ is linearly dependent on the first block and is obtained as a random linear combination of the variables in $\bm{X}^I$ perturbed by additive Gaussian noise. More formally, the dataset is defined as follows:
\begin{itemize}
\item $\bm{X}^I \in \mathbb{R}^{m \times u} : \bm{X}^{I}_{i,j} \sim \mathcal{N}(0,1)$
\item $\bm{X}^D \in \mathbb{R}^{m \times (v-u)} : \bm{X}^D = \bm{X}^{I} \bm{\Phi} + \bm{\mathcal{E}}$
\item $\bm{X} \in \mathbb{R}^{m \times v} : \bm{X} = [\bm{X}^{I} \hspace{2pt} \bm{X}^{D}]$
\end{itemize}
The matrices $\bm{\Phi} \in \mathbb{R}^{u \times (v-u)}$ and $\bm{\mathcal{E}} \in \mathbb{R}^{m \times (v-u)}$ can be generated by sampling each element from a Gaussian distribution, in particular $\bm{\Phi}_{i,j} \sim \mathcal{N}(0,1)$ and $\bm{\mathcal{E}}_{i,j} \sim \mathcal{N}(0,0.1)$. The algorithms were evaluated for the case $u$ = 25, $v$ = 50 and $m$ = 1000.

Table \ref{tab:sim2metrics} reports the values of VE, MI and FP of each algorithm for $k = 5, 10$. All three metrics increase with $k$. For $k$ = 5 the best value of each metric is achieved by the algorithm designed to optimize it, that is, VE, MI and FP achieve their best results with FSCA/L-CA, ITFS and FP-CA, respectively. For $k$ = 10 this pattern is broken for FP, with UFS yielding the lowest FP, although FP-CA is a close second. It is also noteworthy that UFS achieves the same FP as FP-CA for $k$ = 5. This is simply a reflection of the fact that, like FP, the UFS selection process encourages orthogonality among the selected variables. PFS and FOS rank second and third in terms of VE as their selection criterion is related to variance explained. The second best methods in terms of MI are the ones optimizing VE, i.e. FSCA and L-CA. 
\begin{table*}
\centering
\caption{Comparison of the VE, FP and MI values obtained with each algorithm for the `Simulated 2' case study. The best results are highlighted in bold.\\[-0.1in]}
\label{tab:sim2metrics}
\begin{tabularx}{\textwidth}{YYYYY}
\hline
  $|I_S|$  & Algorithm & Variance Explained & Frame Potential & Mutual Information \\ \hline
\multirow{7}*{$k=5$}   & FSCA & \textbf{43.32} & 5.26  & 63.21 \\
& L-CA & \textbf{43.32} & 5.26  & 63.21 \\
& FP-CA & 23.70  & \textbf{5.02}  & 62.06 \\
& PFS & 39.99 & 6.01  & 62.66 \\
& ITFS & 36.18 & 5.21  & \textbf{63.70} \\
& FOS & 39.64 & 6.01  & 61.22 \\
& UFS & 19.30 & \textbf{5.02} & 27.82 \\ \hline
\multirow{7}*{$k=10$}  & FSCA      & \textbf{71.08} & 12.64 & 90.62 \\
& L-CA & \textbf{71.08} & 12.64 & 90.62 \\
& FP-CA & 42.22 & 10.22 & 89.05 \\
& PFS & 68.58 & 12.94 & 90.04 \\
& ITFS & 56.88 & 10.96 & \textbf{91.57} \\
& FOS & 68.45 & 14.19 & 88.59 \\
& UFS & 37.56 & \textbf{10.07} & 54.55 \\ \hline
\end{tabularx}
\end{table*}
According to Table \ref{tab:summary}, in this dataset the best methods in terms of $k_{n\%}$ and $AUC$ are FSCA and L-CA followed closely by PFS and FOS.

\subsection{Pitprops Dataset}
The `Pitprops' dataset was first introduced by \cite{Jeffers} for PCA performance analysis. The data are measurements of physical properties taken from different species of pit props. The scope of that analysis was to determine if the considered pit props were sufficiently strong for deployment in mines. The considered population included samples of different species, size and geographical region. In \cite{Jeffers} the correlation matrix of the original dataset is provided. An approximation of the original 13 attributes is then defined in order to obtain a correlation matrix close to the original one, yielding a data matrix $\bm{X} \in \mathbb{R}^{180\times13}$. The small size of $v$ in this dataset permits an exhaustive search to be performed for the optimal subset of variables for a fixed value of $k$. 

Table \ref{Table2Pitprops} lists the variables selected at each step by each algorithm and compares them with the optimal solution for $|I_S|$ = 7. The number of selected variables that each method has in common with the optimal solution is expressed as the parameter $n_b$ in the final row of the table (i.e. $n_b=|I_S^* \cap I_S|$). ITFS finds the optimum solution. The next closest are UFS with $n_b$ = 5 and FP-CA with $n_b$ = 4. ITFS also yields the highest VE. FSCA, L-CA, PFS and FOS only have 3 variables in common with the optimum subset, but in terms of VE they give the second highest value (85.3\%) followed by FP-CA (83.9\%) and UFS (78.3\%). FSCA/L-CA and FOS select the same seven features, but `length' and `ringbut' are selected as first and third variables, respectively, by FSCA/L-CA, whereas FOS selects them as third and first variables, respectively. Similarly, PFS selects the same variables, but in a different order. Consequently, the VE at $k=7$ for FSCA, L-CA, FOS and PFS is the same. The fact that these four methods achieve a VE that is $98\%$ of the optimum solution with $n_b=3$, while UFS only achieves a VE that is $90\%$ of the optimum with $n_b=5$, highlights how the complex correlation relationships between variables makes optimum variable selection such a challenging problem. 
\begin{table*}
\centering
\caption{First 7 variables of `Pitrops' selected by each algorithm compared with the optimal subset with respect to VE. The order in which the variables are listed reflects the order of their greedy selection. The parameter $n_b$ is the number of selected variables each algorithm has in common with the optimal subset. The best values are highlighted in bold.\\[-0.1in]}
\label{Table2Pitprops}
\begin{tabularx}{\textwidth}{c@{\hskip 0.09in}c@{\hskip 0.09in}c@{\hskip 0.09in}c@{\hskip 0.09in}c@{\hskip 0.09in}c@{\hskip 0.09in}c@{\hskip 0.09in}c}
\hline\noalign{\smallskip}
best & FSCA & L-CA & FP-CA & PFS & ITFS & FOS & UFS\\
\noalign{\smallskip}
\hline
\noalign{\smallskip}
topdiam & length & length & length & length & knots & ringbut & knots\\
testsg & moist & moist & knots & moist & topdiam & moist & bowdist\\ 
ringbut & ringbut & ringbut & diaknot & ringbut & testsg & length & length\\ 
bowdist & clear & clear & clear & bowmax & ringbut & clear & topdiam\\ 
clear & bowmax & bowmax & ovensg & clear & clear & bowmax & ringtop\\ 
knots & ovensg & ovensg & testsg & knots & bowdist & ovensg & clear\\ 
diaknot & knots & knots & bowmax & ovensg & diaknot & knots & diaknot\\ 
\hline
\multicolumn{7}{c}{Variance explained} \\
86.9 & 85.3 & 85.3 & 83.9 & 85.3 & \textbf{86.9} & 85.3 & 78.3\\
\hline
\multicolumn{7}{c}{Number of optimal variables selected ($n_b$)} \\
-& 3 & 3 & 4 & 3 & \textbf{7} & 3 & 5\\
\hline
\end{tabularx}
\end{table*}

\subsection{Wafer Profile Reconstruction}
This case study was provided by a semiconductor manufacturer and is concerned with determining a reduced and optimal set of measurement sites distributed over a silicon wafer surface in order to adequately monitor the uniformity of the thickness of a layer
of material being deposited by a chemical vapour deposition production process. The dataset consists of measurements taken at 50 candidate sites from a set of 316 wafers, hence $\bm{X} \in \mathbb{R}^{316\times50}$. As process monitoring is a time consuming activity, it is desirable to have a reduced set of measurement sites while at the same time maintaining an accurate representation of the surface variation \citep{mcloone2018methodology,susto2019induced}.

%
%
The VE values with increasing $k$ are reported in Table \ref{TableWafer}. While PCA requires 5 components to exceed 99\%, the greedy feature selection algorithms require from a minimum of 7 variables with FSCA, L-CA, PFS, ITFS and FOS to a maximum of 10 with FP-CA. The highest relative performance $r$ is 60\% and is achieved by FSCA and L-CA; ITFS ranks second with 40\%. The $AUC$ values are all similar with just a difference of 0.015 between the highest and the lowest, which indicates that, overall, all the candidate algorithms have a similar trend in terms of VE.
\begin{table*}
\centering
\caption{Number of selected sites $k$ in the `Semiconductor' case study and the corresponding explained variance $V_{\bm{X}}(\hat{\bm{X}}_k)$ for each algorithm. The relative performance ($r$) and $AUC$ metrics are reported at the bottom. The best values after PCA for each $k$ are highlighted in bold.\\[-0.1in]}
\label{TableWafer}
\begin{tabularx}{\textwidth}{c@{\hskip 0.18in}c@{\hskip 0.18in}c@{\hskip 0.18in}c@{\hskip 0.18in}c@{\hskip 0.18in}c@{\hskip 0.18in}c@{\hskip 0.18in}c@{\hskip 0.18in}c}
\hline\noalign{\smallskip}
$k$ & PCA & FSCA  & L-CA & FP-CA & PFS & ITFS & FOS & UFS\\
\noalign{\smallskip}
\hline
\noalign{\smallskip} 
1  & 41.05 & \textbf{38.81} & \textbf{38.81} & \textbf{38.81} & \textbf{38.81} & 35.00    & 37.76 & 16.74 \\
2  & 70.19 & \textbf{67.86} & \textbf{67.86} & 56.50  & \textbf{67.86} & 55.52 & 65.28 & 46.16 \\
3  & 88.35 & \textbf{86.28} & \textbf{86.28} & 82.90  & 86.25 & 72.64 & 85.59 & 62.99 \\
4  & 98.47 & 96.68 & 96.68 & 84.56 & 96.62 & \textbf{96.77} & 96.07 & 94.41 \\
5  & 99.08 & 97.87 & 97.87 & 94.75 & 97.77 & \textbf{98.15} & 97.37 & 97.45 \\
6  & 99.43 & 98.53 & 98.53 & 97.98 & 98.61 & \textbf{98.63} & 98.62 & 98.10 \\
7  & 99.64 & 99.02 & 99.02 & 98.22 & \textbf{99.18} & 99.06 & 99.10  &  98.40 \\
8  & 99.72 & \textbf{99.42} & \textbf{99.42} & 98.57 & 99.37 & \textbf{99.42} & 99.35 & 98.88 \\
9  & 99.79 & \textbf{99.60}  & \textbf{99.60}  & 98.76 & 99.54 & 99.58 & 99.59 & 99.30 \\
10 & 99.85 & \textbf{99.69} & \textbf{99.69} & 99.14 & 99.65 & 99.66 & 99.67 & 99.40 \\
\hline
$r$ & - & \textbf{60} & \textbf{60} & 10 & 30 & 40 & 0 & 0\\
\hline
$AUC$ & 0.979 & \textbf{0.976} & \textbf{0.976} & 0.969 & \textbf{0.976} & 0.970 & 0.975 & 0.961\\
\hline
\end{tabularx}
\end{table*}

The values of VE, FP and MI obtained with each variable selection algorithm are compared in Table \ref{tab:semimetrics} for $k$ = 3, 7. All the three metrics increase with $k$. While the best values of FP and MI are achieved by the corresponding algorithms, the best value of VE for $k = 7$ is given by PFS, followed by FOS and then ITFS. This contrasts with the results for  $k$ = 3 and the results for the `Simulated 2' dataset reported in Table \ref{tab:sim2metrics}, and is a consequence of the high level of redundancy in this dataset with several combinations of $7$ variables sufficient to achieve $99\%$ VE, as evident from Table \ref{SelectionOrderSelectedDataSets}. This is also reflected in the almost identical FP values achieved by all algorithms.  For $k$ = 3, the second ranked algorithm in terms of FP is UFS and in terms of MI it is PFS, while for $k$ = 7 it is FSCA/L-CA and FOS, respectively. Therefore we can see that the algorithms rank quite differently for different metrics, although the overall pattern remains with VE based algorithms performing well in terms of MI, and UFS and FP-CA performing similarly well for FP, but poorly for the other metrics.  
\begin{table*}
\centering
\caption{Comparison of the VE, FP and MI values obtained with each variable selection algorithm for the `Semiconductor' case study. The best results are highlighted in bold.\\[-0.1in]}
\label{tab:semimetrics}
\begin{tabularx}{\textwidth}{YYYYY}
\hline
 $|I_S|$   &  Algorithm    & Variance Explained & Frame Potential & Mutual Information \\ \hline
\multirow{7}*{$k=3$}  & FSCA      & \textbf{86.28} & 8.998     & 96.95 \\
& L-CA & \textbf{86.28} & 8.998     & 96.95 \\
& FP-CA & 82.90  & \textbf{8.997}     & 93.25 \\
& PFS & 86.25 & 8.998     & 97.78 \\
& ITFS & 72.64 & 8.998     & \textbf{97.82} \\
& FOS & 85.59 & 8.998     &  96.22 \\
& UFS & 62.99 &  8.997    & 17.41 \\ \hline
\multirow{7}*{$k=7$}  & FSCA      & 99.02 & 48.982 & 113.30 \\
& L-CA & 99.02 & 48.982 & 113.30 \\
& FP-CA & 98.22 & \textbf{48.979} & 109.87 \\
& PFS & \textbf{99.18} & 48.987 & 114.17 \\
& ITFS & 99.06 & 48.987 & \textbf{115.70} \\
& FOS & 99.10  & 48.987 & 115.31 \\
& UFS & 98.40  & 48.985 & 35.68 \\ \hline
\end{tabularx}
\end{table*}

\subsection{Case Studies from the UCI Repository}
This section considers four datasets taken from the UCI Machine Learning Repository. The datasets are listed and briefly described below. 
\vspace{0.2cm}

\noindent
\textbf{Arrhythmia \citep{guvenir1997supervised}.} This dataset is taken from medical records of arrhythmia patients. It was originally intended to train a classifier able to distinguish between 16 different classes of Arrhythmia defined considering $m$ = 452 patients. The  $v$ = 279 features are both identifiers of the particular patient, e.g. age, sex, weight, and sampled electrocardiogram signal recordings. Since there are some missing values present in the original dataset, we have reduced the number of variables to $v$ = 258. In this scenario, the output of the algorithms $I_S$ can be interpreted as a reduced set of patient data enabling a faster data collection and training process with regard to the classifier development. Additional information on the dataset can be found in \cite{guvenir1997supervised}. \\[+0.1in]
\noindent
\textbf{Sales \citep{tan2014time}.} This dataset is composed of weekly purchase quantities of $m$ = 811 products over $v$ = 52 weeks. The data has a time evolution of 52 weeks and the objective is to identify which weeks to record to get the most accurate prediction of the purchases over the other weeks. For a more detailed dataset description the reader should refer to \cite{tan2015finding}. \\[+0.1in]
\noindent
\textbf{Gases \citep{vergara2012chemical,rodriguez2014calibration}.} This dataset, which was also considered in our previous work \cite{zocco2017mean}, consists of measurements of 8 gas-related parameters provided by each of 16 chemical sensors used to monitor gas level concentrations. Here we considered just the third batch of data, hence $\bm{X} \in \mathbb{R}^{1586\times129}$. \\[+0.1in]
\noindent
\textbf{Music \citep{zhou2014predicting}.} A set of $m$ = 1059 musical tracks from different countries, each one $v$ = 68 samples long, leads to the final dataset $\bm{X} \in \mathbb{R}^{1059\times68}$. Here variable selection identifies a reduced set of samples/columns, each one corresponding to a specific time within the musical tracks. Refer to \cite{zhou2014predicting} for further details.\\[-0.1in]

The results for the UCI case studies are summarized in Table \ref{FinalTable} for four different compression thresholds $k_{n\%}$. The best performing algorithm is highlighted in bold in each case. Overall, FSCA/L-CA achieves the highest compression performance with 12 entries in bold followed by PFS with 9, and then FOS and ITFS with 3 each. The `Music' dataset is the most difficult to compress; for a reconstruction accuracy of 80\% the number of samples to record is reduced by 59\% with the best algorithm, but it is only reduced by 29\% when the desired reconstruction accuracy is 95\%. `Sales' only requires a large number of variables at the 99\% accuracy level. For lower threshold levels, data from fewer than 10\% of the weeks is sufficient to predict the purchases for the remaining weeks using FSCA, L-CA, PFS or ITFS. In the case of `Arrhythmia' the best performing algorithm achieves a 72\% reduction in the number of variables to reach 99\% reconstruction accuracy, while in the case of `Gases' the reduction is a remarkable 97\%.

\begin{table*}
	\centering
	\caption{Variable selection algorithm performance on selected UCI benchmark datasets. Four metrics are presented for each algorithm: $k_{80\%}$, $k_{90\%}$, $k_{95\%}$ and $k_{99\%}$. The best values are highlighted in bold.\\[-0.1in]}
	\label{FinalTable}
	\begin{tabularx}{\textwidth}{c@{\hskip 0.15in}c@{\hskip 0.15in}c@{\hskip 0.15in}c@{\hskip 0.15in}c@{\hskip 0.15in}c@{\hskip 0.15in}c@{\hskip 0.15in}c@{\hskip 0.15in}c}
		\hline
		Dataset                               & $k_{n\%}$ & FSCA & L-CA & FP-CA & PFS & ITFS  & FOS & UFS                                                        \\
		\hline
		\multirow{4}*{\makecell{Arrhyt. \\ \scriptsize(452 $\times$ 258)}}       
		& $k_{80\%}$ & \textbf{20} & \textbf{20} & 140 & 21 & 78  & 68  & 119 \\
		& $k_{90\%}$ & \textbf{33} & \textbf{33} & 190 & 34 & 118 & 122 & 153 \\
		& $k_{95\%}$ & \textbf{47} & \textbf{47} & 241 & \textbf{47} & 144 & 178 & 193 \\
		& $k_{99\%}$ & \textbf{71} & \textbf{71} & 256 & \textbf{71} & 173 & 240 & 239 \\
		\hline
		\multirow{4}*{\makecell{Sales \\ \scriptsize(812 $\times$ 52)}} 
		& $k_{80\%}$ & 1 & 1 & 1 & 1 & 1 & 1 & 1  \\
		& $k_{90\%}$ & 1  & 1  & 1  & 1  & 1  & 1  & 1 \\
		& $k_{95\%}$ & \textbf{5}  & \textbf{5}  & 7  & \textbf{5} & \textbf{5} & 9  & 13 \\
		& $k_{99\%}$ & \textbf{38} & \textbf{38} & 41 & 39 & 40 & 42 & 42 \\  
		\hline
		\multirow{4}*{\makecell{Gases \\ \scriptsize(1586 $\times$ 129)}}  
		& $k_{80\%}$ & \textbf{1} & \textbf{1} & \textbf{1} & \textbf{1} & 2  & 2 & 2 \\
		& $k_{90\%}$ & 2 & 2 & 2 & 2 & 2  & 2 & 2  \\
		& $k_{95\%}$  & \textbf{2} & \textbf{2} & \textbf{2} & \textbf{2} & \textbf{2}  & 3 & 4 \\
		& $k_{99\%}$ & \textbf{3} & \textbf{3} & 6 & \textbf{3} & 11 & 9 & 17  \\
		\hline
		\multirow{4}*{\makecell{Music \\ \scriptsize(1059 $\times$ 68)}}  
		& $k_{80\%}$ &\textbf{28} & \textbf{28} & 29 & 29 & 31 & \textbf{28} & 39 \\
		& $k_{90\%}$& \textbf{40} & \textbf{40} & 42 & \textbf{40} & 41 & \textbf{40} & 49  \\
		& $k_{95\%}$ & 49 & 49 & 53 & \textbf{48} & 49 & 49 & 56 \\
		& $k_{99\%}$ & \textbf{61} & \textbf{61} & 64 & \textbf{61} & \textbf{61} & \textbf{61} & 64 \\
		\hline
	\end{tabularx}
\end{table*}


\subsection{\hlight{Larger Datasets}}
\hlight{This section presents results for the four larger datasets listed in Table \ref{tab:datasetsSummary}, namely, PlasmaEtch, YaleB, WaferIPCM and USPS. The \textbf{PlasmaEtch} dataset is derived from 2046 channel Optical Emission Spectrum (OES) measurements recorded from a semiconductor manufacturing plasma etch process during the processing of 2194 wafers \citep{puggini2018_anomaly}. Each variable in the dataset is the mean value of light intensity time series data recorded for a specific wavelength (channel) during a trench etching step.  \textbf{YaleB}$^6$ is a subset of the extended Yale Face Database B \citep{georghiades2001_YaleB} containing 2414 cropped $32 \times 32$ pixel face images of 38 individuals under different lighting conditions.   \textbf{WaferIPCM}$^5$ is a microelectronics fabrication process monitoring dataset consisting of 152 inline process control measurements per wafer recorded for 7164 wafers \citep{olszewski2001generalized}. \textbf{USPS}$^6$ is a subset of the USPS handwritten digit database \citep{hull1994USPS} consisting of 9298 $16 \times 16$ handwritten digit images.	The, `PlasmaEtch' and `YaleB' case studies have the largest number of features $v$, $2046$ and $1024$, respectively, and a similar value of $m$, whereas the other two datasets have many more measurements than variables (i.e. $m >> v$).
	
Figure \ref{fig:figuresLargerDatasets} shows the VE performance of the variable selection algorithms on these datasets for the first 10 selected variables in each case. It is evident that `PlasmaEtch' is a highly correlated dataset, with almost 100\% VE achieved with only 10 of its 2046 variables. In contrast, USPS is the least correlated dataset with 10 variables only able to achieve 63\% VE.  
Overall, algorithm performances are similar to that observed with the smaller datasets with FSCA and L-CA yielding almost identical results, followed closely by PFS and then FOS. The only significant deviation with PFS is for $k=2$ in `YaleB', where its VE is 6.5\% lower than that achieved by FSCA. FOS yields identical results to FSCA on `YaleB' and WaferIPCM, but is inferior to FSCA on the other two datasets.  Again, UFS and FP-CA are the worst performing algorithms.  While ITFS achieves similar performance to FSCA for $k>3$ on `YaleB', it lags behind FSCA on the other datasets, and on `PlasmaEtch' in particular, where it is among the worst performing algorithms.   
\begin{figure}
\begin{subfigure}{0.5\textwidth}
\includegraphics[width=\textwidth]{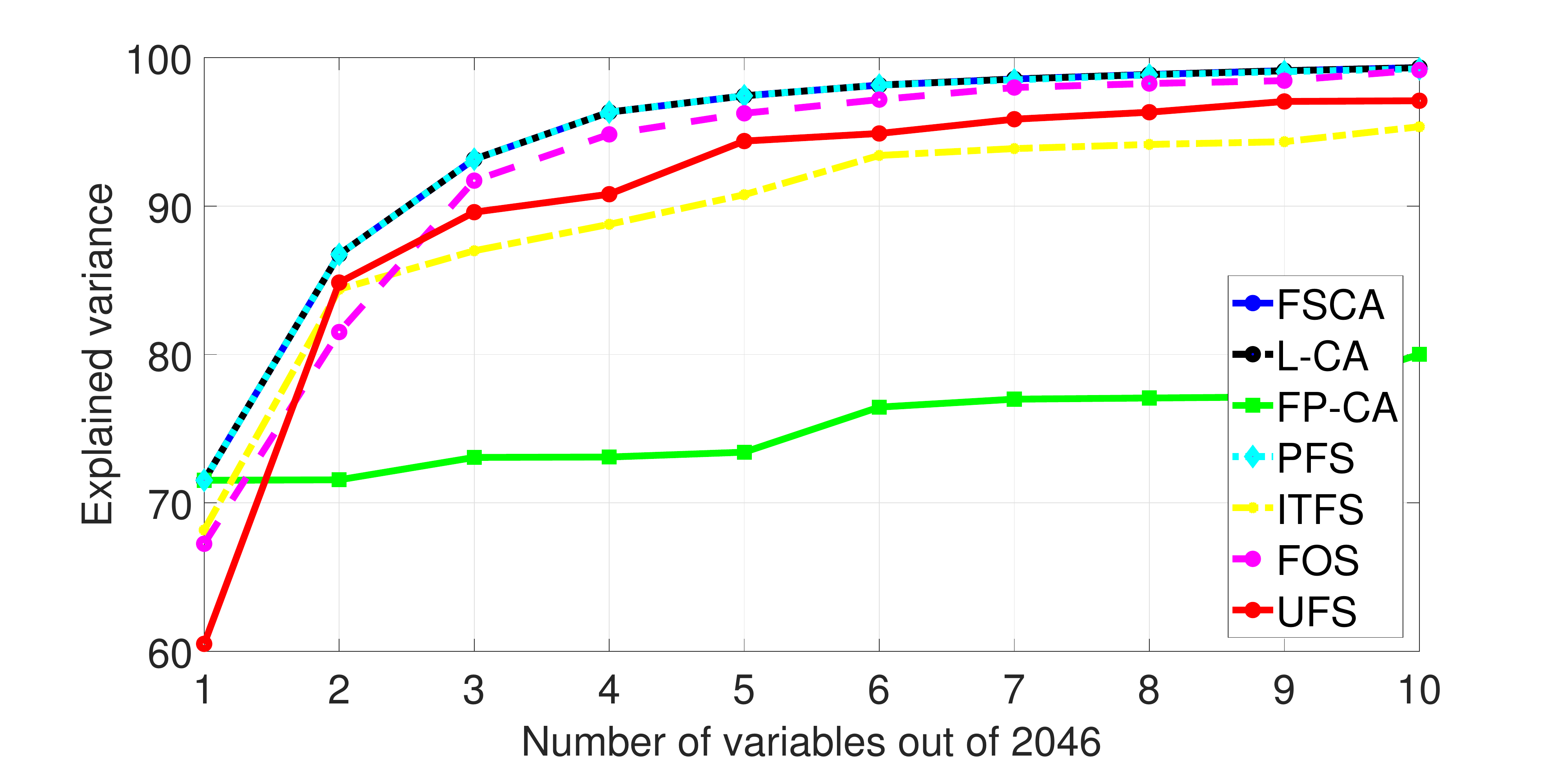} 
\caption{PlasmaEtch}
\end{subfigure}
\begin{subfigure}{0.5\textwidth}
\includegraphics[width=\textwidth]{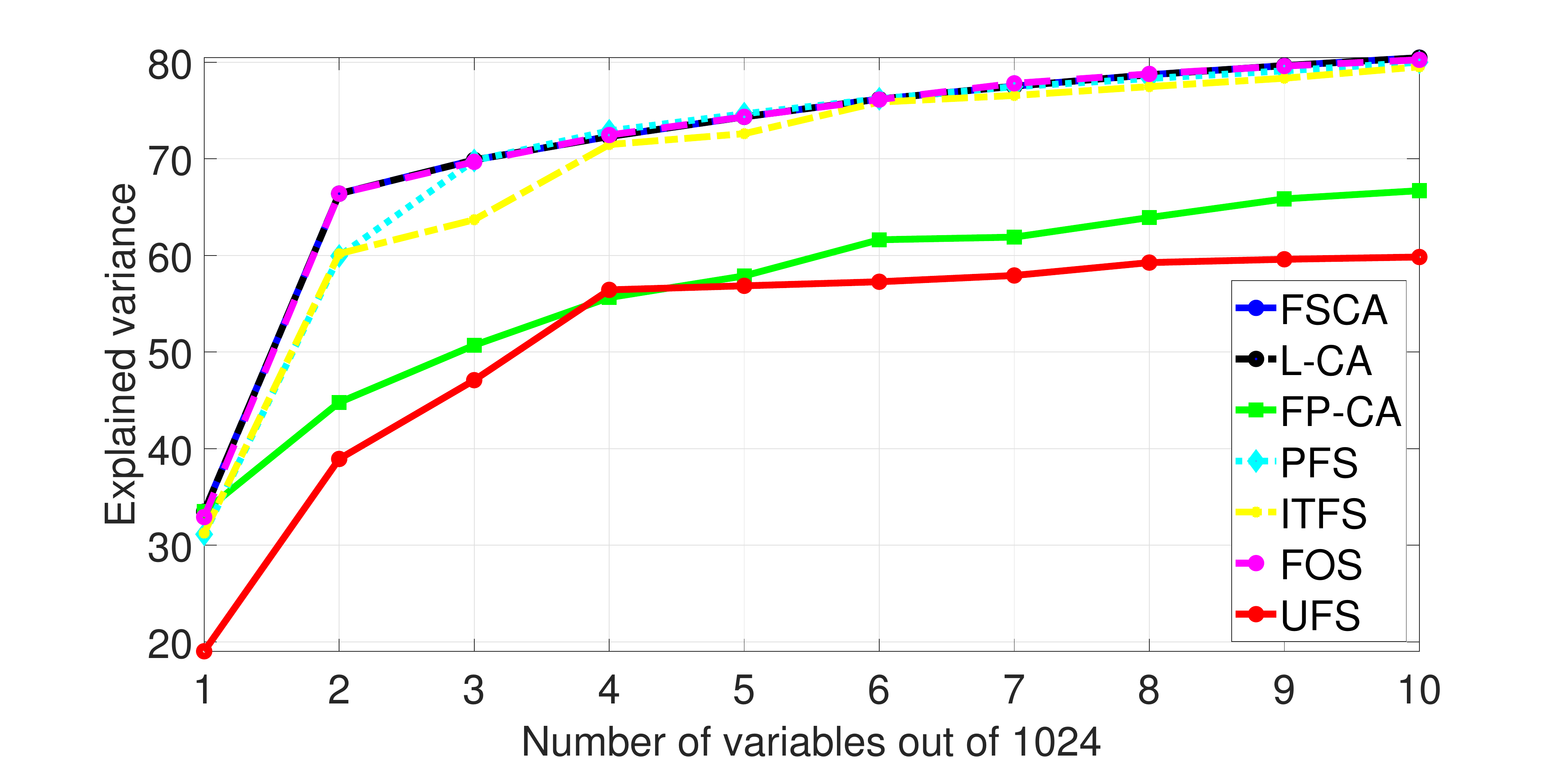}
\caption{YaleB}
\end{subfigure}
\begin{subfigure}{0.5\textwidth} 
\includegraphics[width=\textwidth]{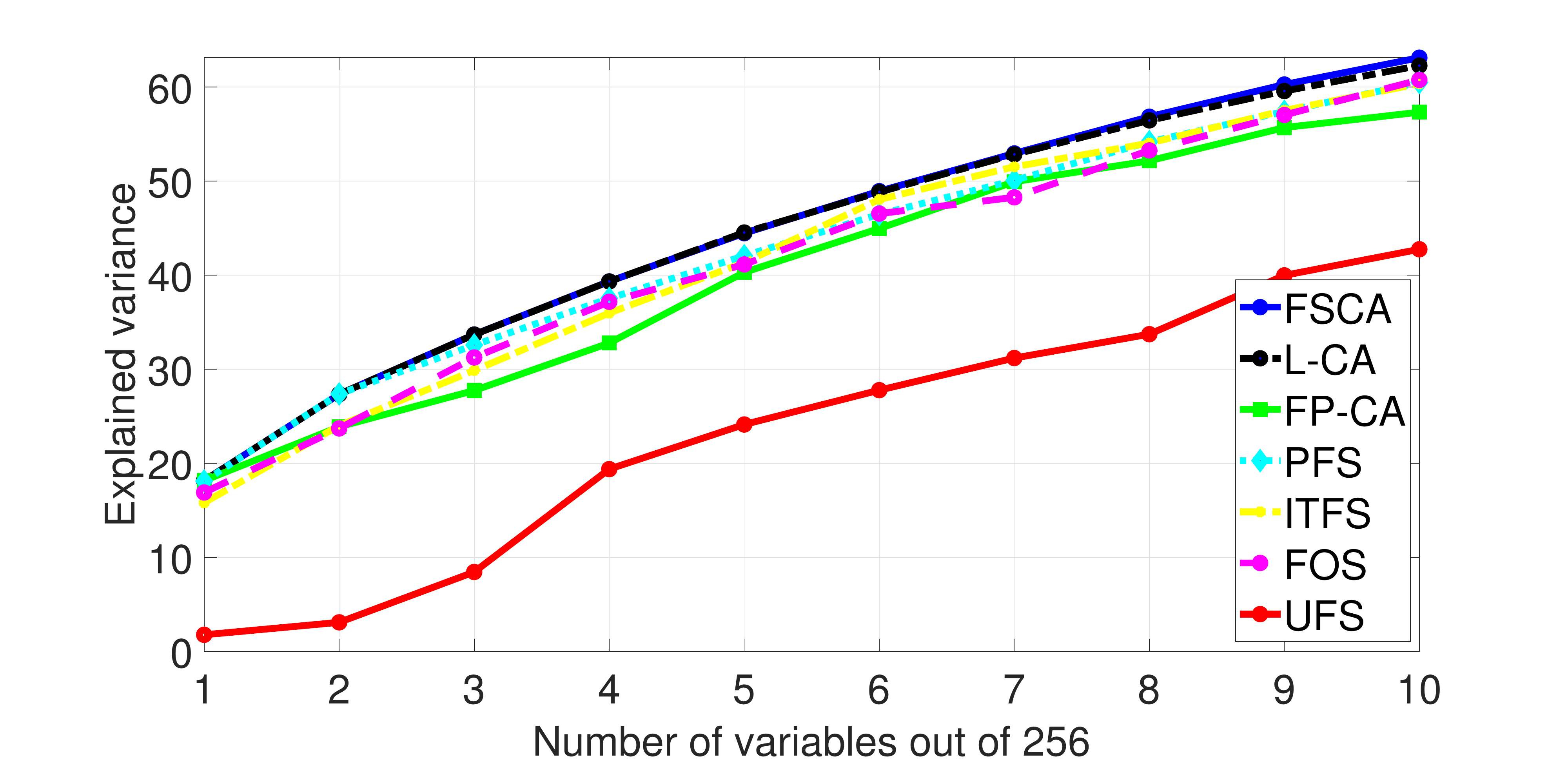}
\caption{USPS}
\end{subfigure}
\begin{subfigure}{0.5\textwidth}
\includegraphics[width=\textwidth]{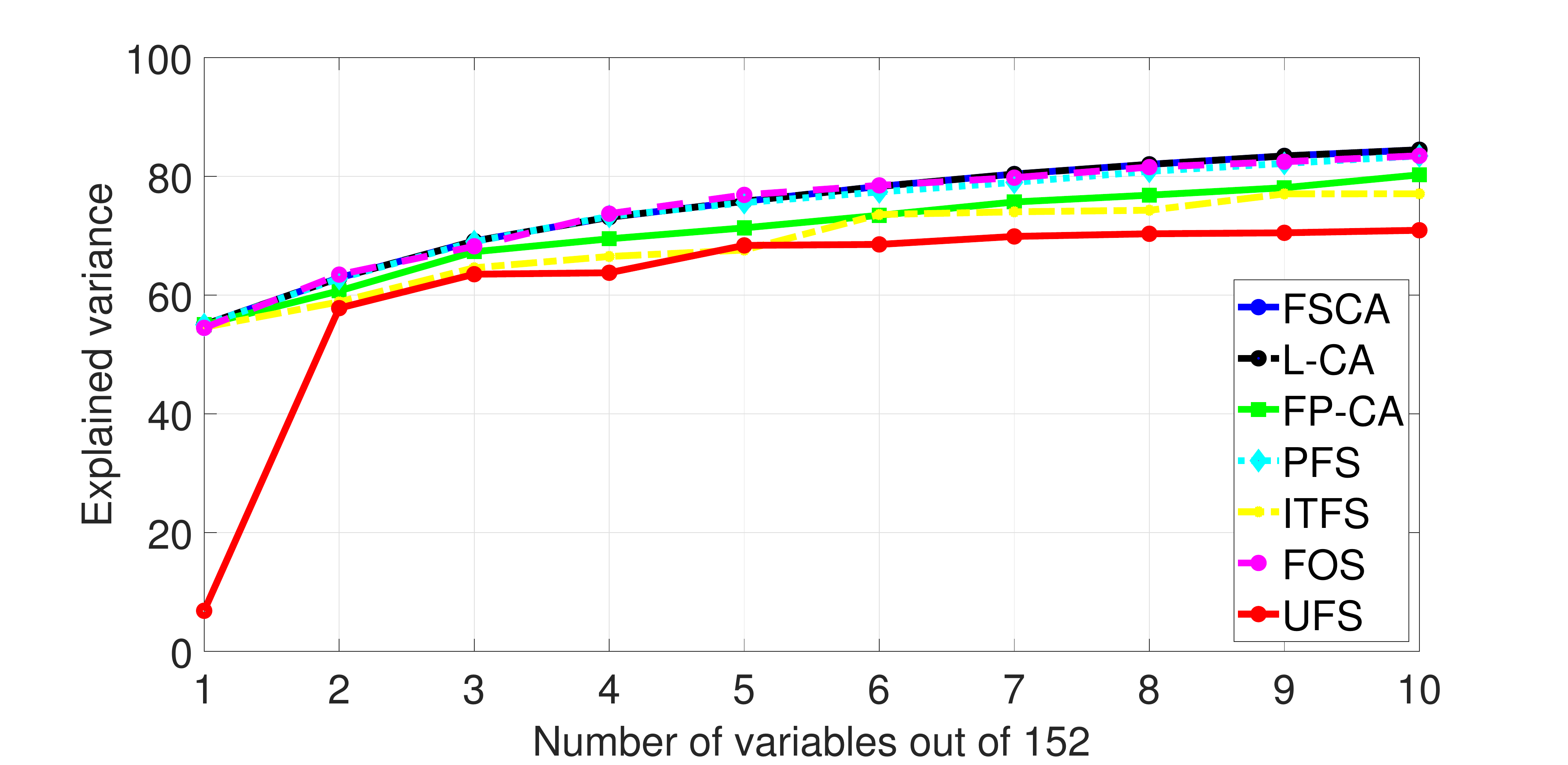}
\caption{WaferIPCM}
\end{subfigure}
\caption{VE as a function of the number of selected variables $k$ for the larger datasets.}
\label{fig:figuresLargerDatasets}
\end{figure}
}

\newpage
\subsection{\hlight{Computation Times}}
\hlight{To assess the efficiency of L-FSCA relative to FSCA and provide a comparison of the computation requirements of all algorithms considered in the paper, in this section two tables of timing results are presented. In Table \ref{tab:L-FSCAvFSCA}, the speed-up achieved by L-FSCA over FSCA is computed for a range of dataset dimensions ($m \times v$) and $k$ values for: (1) randomly generated data, $\bm{X}_{i,j} \sim \mathcal{N}(0,1)$; (2) data with the same correlation structure as the 'Simulated dataset 2' case study, but with $u=50$. In Table \ref{tab:ComputTimesLargerDatasets}, the speed-up ratio relative to FSCA is reported for each algorithm for the four largest datasets for selected values of $k$. Results are also included for the `Sim. 2 ($u=50$)' and `Random' datasets with $m=2194$ and $v=2046$. The corresponding FSCA execution times (in seconds) are also recorded.
	
The speed-up ratio, $s_i$, for the $i$th algorithm is defined as $s_i=\frac{t_{\text{FSCA}}}{t_{i}}$, where $t_i$ is the execution time of the $i$th algorithm and $t_{\text{FSCA}}$ is the execution time of FSCA. If $s_i<1$ then the algorithm is slower than FSCA, in which case the values have been reported as a fraction $1/q_i$, where $q_i$ indicates how many times slower algorithm $i$ is than FSCA.
	
	\begin{table*}
		\centering
		\caption{\hlight{L-FSCA computation time relative to FSCA. The results presented are the median speed-up relative to FSCA (based on 10 Monte Carlo simulations) for different problem dimensions ($m \times v$) for: (1) randomly generated data; (2) data from the `Simulated 2' case study with $u=50$.\\[-0.1in]}}
		\label{tab:L-FSCAvFSCA}
		\begin{tabularx}{0.99\textwidth}{c@{\hskip 0.25in}c@{\hskip 0.25in}c@{\hskip 0.25in}c@{\hskip 0.25in}c@{\hskip 0.25in}c@{\hskip 0.25in}c@{\hskip 0.25in}c}
			\hline
			\multirow{2}{*}{\makecell{Dataset \\[0.1in]  }}     & $m$    & 500  & 500  & 500  & 5000 & 5000 & 5000  \\[-0.1in]
			& $v$    & 100  & 200  & 400  & 100  & 200  & 400   \\
			\hline
			\multirow{4}{*}{\makecell{Random \\$\bm{X}_{i,j} \sim \mathcal{N}(0,1)$}}         & $k=5~$  & 2.56 & 2.95 & 2.24 & 1.79 & 2.78 & 3.97  \\
			& $k=10$ & 2.18 & 3.14 & 3.06 & 2.90 & 3.88 & 6.51  \\
			& $k=20$ & 2.60 & 2.93 & 3.85 & 3.36 & 4.80 & 9.43  \\
			& $k=50$ & 2.76 & 3.18 & 3.97 & 3.95 & 5.43 & 11.49 \\
			\hline
			\multirow{4}{*}{\makecell{Sim. 2\\($u=50$)}} & $k=5~$  & 2.31 & 2.93 & 2.59 & 1.90 & 2.70 & 3.78  \\
			& $k=10$ & 2.36 & 2.56 & 2.99 & 2.30 & 3.44 & 5.67  \\
			& $k=20$ & 2.21 & 2.53 & 2.76 & 2.69 & 3.73 & 6.40  \\
			& $k=50$ & 2.34 & 2.54 & 3.24 & 2.86 & 3.63 & 5.38 \\
			\hline
		\end{tabularx}
	\end{table*}

\begin{table*}
	\centering
	\caption{\hlight{Computation times relative to FSCA for each algorithm for the larger datasets and for `Sim. 2 ($u=50$)' and `Random' datasets  of the same dimension as `PlasmaEtch'.  The results presented are the median speed-up relative to FSCA (based on 10 Monte Carlo simulations). The best results are highlighted in bold.\\[+0.1in]}}
	\label{tab:ComputTimesLargerDatasets}
	\begin{tabularx}{1.03\textwidth}{c@{\hskip 0.13in}c@{\hskip 0.13in}c@{\hskip 0.15in}c@{\hskip 0.15in}c@{\hskip 0.15in}c@{\hskip 0.15in}c@{\hskip 0.15in}c@{\hskip 0.15in}c}
		\hline
		\makecell{Dataset \\[-0.1in] \scriptsize(m $\times$ v)} 
		& $k$ & \makecell{ \scriptsize  FSCA \\[-0.1in] \scriptsize time(s)}  & \makecell{L-CA} & \makecell{FP-CA} & \makecell{PFS} & \makecell{ITFS} & \makecell{FOS} & \makecell{UFS}\\
		\noalign{\smallskip}
		\hline
		\multirow{3}*{\makecell{WaferIPCM \\ \scriptsize(7164 $\times$ 152)}} & 5 
		& 0.069 & $\bm{1.44}$ & 1/53.0 & 1.10 & 1/121.9 & 1/22.5 & 1.069 \\
		&10 & 0.124 & $\bm{1.82}$ & 1/65.5 & 1.043 & 1/129.2 & 1/28.9 & 1/1.8\\
		&20 & 0.271 & $\bm{2.93}$ & 1/66.9 & 1/1.1 & 1/112.7 & 1/31.3 & 1/3.5\\
		\hline
		\noalign{\smallskip}
		\multirow{3}*{\makecell{USPS \\ \scriptsize(9298 $\times$ 256)}} 
		& 5 & 0.427 & 2.09 & 1/36.6 & 1.33 & 1/74.4 & 1/13.0 & $\bm{3.03}$\\
		&10 & 0.840 & $\bm{2.57}$ & 1/42.4 & 1.14 & 1/74.3 & 1/15.2 & 1.65\\
		&20 & 1.687 & $\bm{3.41}$ & 1/47.5 & 1.05 & 1/72.2 & 1/16.5 & 1/1.2\\
		\hline
		\multirow{3}*{\makecell{YaleB \\ \scriptsize(2414 $\times$ 1024)}} 
		& 5 & 1.565 & 2.10 & 1/53.0 & 3.59 & 1/165.5 & 1/14.0 & $\bm{15.21}$ \\
		&10 & 3.038 & 2.90 & 1/58.7 & 3.64 & 1/169.6 & 1/16.0 & $\bm{ 9.89}$ \\
		&20 & 6.040 & 4.22 & 1/62.8 & 3.13 & 1/168.7 & 1/17.2 & $\bm{5.79}$ \\
		\hline
		\multirow{3}*{\makecell{PlasmaEtch \\ \scriptsize(2194 $\times$ 2046)}} 
		& 5 &  7.745 & 1.29 & 1/60.4 & 29.12 & 1/231.3 & 1/10.1 & $\bm{35.52}$ \\
		&10 & 15.554 & 1.79 & 1/68.9 & 23.66 & 1/230.0 & 1/11.4 & $\bm{23.55}$\\
		&20 & 30.939 & 2.35 & 1/73.8 & 15.93 & 1/234.2 & 1/12.2 & $\bm{17.38}$\\
		\hline
		\multirow{3}*{\makecell{Sim. 2 \\[-0.1in]($u=50$) \\ \scriptsize(2194 $\times$ 2046)}}
		& 5 &  7.934 &  4.73 & 1/60.2 & 2.94 & 1/224.1 &  1/8.8 & $\bm{26.90}$ \\
		&10 & 16.026 &  8.87 & 1/67.3 & 1.65 & 1/215.7 &  1/9.9 & $\bm{28.33}$\\
		&20 & 30.395 & 12.64 & 1/73.3 & 2.23 & 1/222.2 & 1/10.9 & $\bm{18.02}$\\
		\hline
		\multirow{3}*{\makecell{Random \\ \scriptsize(2194 $\times$ 2046)}} 
		& 5 & 7.784 & 4.82 & 1/63.5 & 1.99 & 1/235.4 & 1/10.3 & $\bm{32.80}$ \\
		&10 & 16.283 & 9.23 & 1/68.5 & 1.84 & 1/228.3 & 1/10.7 & $\bm{24.35}$\\
		&20 & 31.697 & $\bm{17.01}$ & 1/73.1 & 1.85 & 1/229.1 & 1/11.2 & 16.15\\
		\hline
	\end{tabularx}
\end{table*}

 The results in Table \ref{tab:L-FSCAvFSCA} and Table \ref{tab:ComputTimesLargerDatasets} show that the speed-up achievable with L-FSCA (L-CA) varies considerably depending on the problem dimension, value of $k$, and the correlation structure in the data. The lowest speed-up observed is $1.29$ with the highly correlated `PlasmaEtch' dataset ($k=5$) and the largest speed-up is $17.01$ with the $2194 \times 2046$ randomly generated data ($k=20$). In general the larger the problem dimension ($m \times v$), the larger the value of $k$, and the less correlated the data, the greater the speed-up achievable with L-FSCA.
 
 It is interesting to note that PFS proves to be exceptionally efficient to compute for `PlasmaEtch' (16$-$30 times faster than FSCA). This contrasts with L-FSCA which only achieves modest speed-ups for this problem (1.3$-$2.3). These differences are largely due to the highly correlated nature of this problem, which results in: (1) a covariance matrix spectral distribution that favours rapid convergence of the NIPALS procedure; (2) a reduction in the efficiency of the lazy greedy procedure due to the need to search through more of the ordered list of marginal gains before locating the next best variable.  As evidence of this, when PFS is applied to `Sim. 2' and `Random' datasets of the same dimension as `PlasmaEtch' it is only 1.7$-$2.9 times faster than FSCA while L-FSCA  is 4.7$-$17.0 times faster. Hence, it can be concluded that problem characteristics have a big impact on the computational efficiency of both algorithms. 
 
 UFS is the most scalable of the variable selection algorithms with respect to problem dimension ($m \times v$). It is between 17$-$36 times faster to compute than FSCA for `PlasmaEtch' and 5-16 times faster for `YaleB'.  However, its performance deteriorates with increasing $k$, which contrasts with L-FSCA which becomes more competitive with increasing $k$. 
 	
 As expected, FP-CA and ITFS are the most computationally expensive algorithms and quickly become prohibitive to compute as the problem size increases. They are 36$-$74 and 72$-$236 times slower that FSCA, respectively, for the case studies considered. FOS is also at least an order of magnitude slower than FSCA for most problems.
 
 It should be noted that the execution time of an algorithm depends on how it is implemented (in contrast with the Big-O complexity in Table \ref{TableComplexity}). Hence, when considering the results in Table \ref{tab:ComputTimesLargerDatasets}, the reader should bear in mind that FSCA, L-FSCA and PFS have been implemented efficiently leveraging the work of \cite{IEEEPuggini}, whereas the implementations of the other algorithms are as reported in the literature and have not undergone the same level of code optimisation.   
 
}

\section{Conclusions} \label{concl}
\hlight{This paper has proposed a novel \emph{greedy unsupervised} variable selection algorithm, L-FSCA, and compared its performance with six other variable selection algorithms, five of which, FSCA, PFS, ITFS, FOS-MOD and UFS, are taken from the literature and the sixth, FSFP-FSCA, is an enhanced implementation of FSFP, also taken from the literature. The comparisons are based on two simulated datasets and ten real world case studies.

The development of L-FSCA was motivated by an assumption that while variance explained is not a submodular function, it is sufficiently close to being submodular in practice to warrant exploitation of an efficient \emph{lazy} greedy implementation that is valid for such functions. The experimental evidence confirms the validity of this assumption with L-FSCA yielding almost identical accuracy to FSCA for the broad range of case studies investigated, while being substantially faster to execute for most problems. L-FSCA yielded a reduction in computation time of between 22\% and 94\% for the problems considered with speed-ups achievable increasing with increasing problem size, $m \times v$, and number of selected variables, $k$. 

The experimental comparison of the different variable selection algorithms showed that, with respect to the VE performance metric, the algorithms based on variance explained or the closely related squared correlation selection functions substantially outperform the mutual information based algorithm (ITFS), and the algorithms employing functions that encourage orthogonality (FSFP-FSCA and UFS). They are also generally competitive with these other methods with respect to their native metrics, i.e. mutual information (MI) and frame potential (FP). Overall, L-FSCA/FSCA is the best performing algorithm across all case studies with regard to achieving the greatest data compression, as reflected in the AUC metric, followed closely by PFS. UFS and FSFP-FSCA are consistently the worst performing methods. }

\hlight{In terms of Big-O complexity, the most efficient algorithms among those considered are UFS, PFS (with principal components computed via NIPALS) and L-FSCA. UFS is the least sensitive to problem dimension, $v$, making it much more efficient to compute than the other algorithms when $v$ is large. However, its computational advantages diminish with increasing $m$ and $k$, with the result that it is outperformed by the other algorithms when $m>>v$, particularly for larger values of $k$ (see, for example, the `WaferIPCM' case study). 
	
The relative performance of PFS and L-FSCA is problem dependent with PFS faster than L-FSCA for 5 of the 18 dataset-variable selection combinations investigated in Table \ref{tab:ComputTimesLargerDatasets}, with the biggest differences occurring for the highly correlated `PlasmaEtch' case study.  This arises because the complexity of PFS is $O(Nkmv)$, with $N$ determined by the implementation of NIPALS and the spectral characteristics of the data covariance matrix, while the complexity of L-FSCA tends towards $O(kmv)$ as a function of the efficiency of the lazy search, both of which are impacted by the level of correlation in the data. High levels of correlation among variables is detrimental to the efficiency of L-FSCA, but beneficial to the rate of convergence of the NIPALS procedure in PFS.

The main limitation that arises with L-FSCA is that its computation time is problem dependent and therefore cannot be determined a priori, a characteristic it has in common with PFS, its closest competitor. However, unlike PFS, since it is lower bounded by the computational complexity of FSCA, it is guaranteed to be at least as fast as FSCA, whose computation time is deterministic. Also, as already noted, L-FSCA does not enjoy theoretical performance guarantees because VE does not meet the submodularity requirements that underpin the available theory, a limitation its shares with FSCA, PFS and FOS.  However, this in no way diminishes its practical value and effectiveness, as evident from its excellent performance over the broad range of case studies investigated.  Thus, considering overall computational efficiency, VE performance, and simplicity of use, L-FSCA is the algorithm of choice for unsupervised variable selection.}



\section*{Acknowledgements}
The first author gratefully acknowledges the financial support provided by Irish Manufacturing Research (IMR) for this research. \hlight{All the authors thank the anonymous reviewers for their valuable comments, and the researchers who contributed the publicly available datasets as noted in Table \ref{tab:datasetsSummary}}.

\bibliography{References}

\begin{thebibliography}{52}
\expandafter\ifx\csname natexlab\endcsname\relax\def\natexlab#1{#1}\fi
\providecommand{\url}[1]{\texttt{#1}}
\providecommand{\href}[2]{#2}
\providecommand{\path}[1]{#1}
\providecommand{\DOIprefix}{doi:}
\providecommand{\ArXivprefix}{arXiv:}
\providecommand{\URLprefix}{URL: }
\providecommand{\Pubmedprefix}{pmid:}
\providecommand{\doi}[1]{\href{http://dx.doi.org/#1}{\path{#1}}}
\providecommand{\Pubmed}[1]{\href{pmid:#1}{\path{#1}}}
\providecommand{\bibinfo}[2]{#2}
\ifx\xfnm\relax \def\xfnm[#1]{\unskip,\space#1}\fi
\bibitem[{Bendel and Afifi(1977)}]{bendel1977comparison}
\bibinfo{author}{Bendel, R.B.}, \bibinfo{author}{Afifi, A.A.},
  \bibinfo{year}{1977}.
\newblock \bibinfo{title}{Comparison of stopping rules in forward
  “stepwise” regression}.
\newblock \bibinfo{journal}{Journal of the American Statistical Association}
  \bibinfo{volume}{72}, \bibinfo{pages}{46--53}.
\bibitem[{Bian et~al.(2017)Bian, Buhmann, Krause and
  Tschiatschek}]{bian2017guarantees}
\bibinfo{author}{Bian, A.A.}, \bibinfo{author}{Buhmann, J.M.},
  \bibinfo{author}{Krause, A.}, \bibinfo{author}{Tschiatschek, S.},
  \bibinfo{year}{2017}.
\newblock \bibinfo{title}{Guarantees for greedy maximization of non-submodular
  functions with applications}, in: \bibinfo{booktitle}{International
  Conference on Machine Learning}, \bibinfo{organization}{PMLR}. pp.
  \bibinfo{pages}{498--507}.
\bibitem[{Chepuri and Leus(2015)}]{chepuri2015sparsity}
\bibinfo{author}{Chepuri, S.P.}, \bibinfo{author}{Leus, G.},
  \bibinfo{year}{2015}.
\newblock \bibinfo{title}{Sparsity-promoting sensor selection for non-linear
  measurement models}.
\newblock \bibinfo{journal}{IEEE Transactions on Signal Processing}
  \bibinfo{volume}{63}, \bibinfo{pages}{684--698}.
\bibitem[{Conforti and Cornu{\'e}jols(1984)}]{conforti1984}
\bibinfo{author}{Conforti, M.}, \bibinfo{author}{Cornu{\'e}jols, G.},
  \bibinfo{year}{1984}.
\newblock \bibinfo{title}{Submodular set functions, matroids and the greedy
  algorithm: tight worst-case bounds and some generalizations of the
  rado-edmonds theorem}.
\newblock \bibinfo{journal}{Discrete Applied Mathematics} \bibinfo{volume}{7},
  \bibinfo{pages}{251--274}.
\bibitem[{Cui and Dy(2008)}]{cui2008orthogonal}
\bibinfo{author}{Cui, Y.}, \bibinfo{author}{Dy, J.G.}, \bibinfo{year}{2008}.
\newblock \bibinfo{title}{Orthogonal principal feature selection}, in:
  \bibinfo{booktitle}{The Sparse Optimization and Variable Selection Workshop
  at the 25th International Conference on Machine Learning},
  \bibinfo{organization}{Helsinki, Finland}.
\bibitem[{Das and Kempe(2008)}]{das2008algorithms}
\bibinfo{author}{Das, A.}, \bibinfo{author}{Kempe, D.}, \bibinfo{year}{2008}.
\newblock \bibinfo{title}{Algorithms for subset selection in linear
  regression}, in: \bibinfo{booktitle}{Proceedings of the Fortieth Annual ACM
  Symposium on Theory of Computing}, \bibinfo{publisher}{ACM},
  \bibinfo{address}{New York, NY, USA}. p. \bibinfo{pages}{45–54}.
\bibitem[{Das and Kempe(2011)}]{das2011submodular}
\bibinfo{author}{Das, A.}, \bibinfo{author}{Kempe, D.}, \bibinfo{year}{2011}.
\newblock \bibinfo{title}{Submodular meets spectral: Greedy algorithms for
  subset selection, sparse approximation and dictionary selection}, in:
  \bibinfo{booktitle}{Proceedings of the 28th International Conference on
  International Conference on Machine Learning},
  \bibinfo{publisher}{Omnipress}, \bibinfo{address}{Madison, WI, USA}. p.
  \bibinfo{pages}{1057–1064}.
\bibitem[{Das and Kempe(2018)}]{das2018approximate}
\bibinfo{author}{Das, A.}, \bibinfo{author}{Kempe, D.}, \bibinfo{year}{2018}.
\newblock \bibinfo{title}{Approximate submodularity and its applications:
  {Subset} selection, sparse approximation and dictionary selection}.
\newblock \bibinfo{journal}{The Journal of Machine Learning Research}
  \bibinfo{volume}{19}, \bibinfo{pages}{74--107}.
\bibitem[{d'Aspremont et~al.(2005)d'Aspremont, Ghaoui, Jordan and
  Lanckriet}]{d2005direct}
\bibinfo{author}{d'Aspremont, A.}, \bibinfo{author}{Ghaoui, L.E.},
  \bibinfo{author}{Jordan, M.I.}, \bibinfo{author}{Lanckriet, G.R.},
  \bibinfo{year}{2005}.
\newblock \bibinfo{title}{A direct formulation for sparse {PCA} using
  semidefinite programming}, in: \bibinfo{booktitle}{Advances in Neural
  Information Processing Systems}, pp. \bibinfo{pages}{41--48}.
\bibitem[{Flynn and McLoone(2011)}]{flynn2011max}
\bibinfo{author}{Flynn, B.}, \bibinfo{author}{McLoone, S.},
  \bibinfo{year}{2011}.
\newblock \bibinfo{title}{Max separation clustering for feature extraction from
  optical emission spectroscopy data}.
\newblock \bibinfo{journal}{IEEE Transactions on Semiconductor Manufacturing}
  \bibinfo{volume}{24}, \bibinfo{pages}{480--488}.
\bibitem[{Georghiades et~al.(2001)Georghiades, Belhumeur and
  Kriegman}]{georghiades2001_YaleB}
\bibinfo{author}{Georghiades, A.}, \bibinfo{author}{Belhumeur, P.},
  \bibinfo{author}{Kriegman, D.}, \bibinfo{year}{2001}.
\newblock \bibinfo{title}{From few to many: Illumination cone models for face
  recognition under variable lighting and pose}.
\newblock \bibinfo{journal}{IEEE Transactions on Pattern Analysis and Machine
  Intelligence} \bibinfo{volume}{23}, \bibinfo{pages}{643--660}.
\bibitem[{Guvenir et~al.(1997)Guvenir, Acar, Demiroz and
  Cekin}]{guvenir1997supervised}
\bibinfo{author}{Guvenir, H.A.}, \bibinfo{author}{Acar, B.},
  \bibinfo{author}{Demiroz, G.}, \bibinfo{author}{Cekin, A.},
  \bibinfo{year}{1997}.
\newblock \bibinfo{title}{A supervised machine learning algorithm for
  arrhythmia analysis}, in: \bibinfo{booktitle}{Computers in Cardiology}, pp.
  \bibinfo{pages}{433--436}.
\bibitem[{Han et~al.(2018)Han, Wang, Zhang, Li and Xu}]{Han2018}
\bibinfo{author}{Han, K.}, \bibinfo{author}{Wang, Y.}, \bibinfo{author}{Zhang,
  C.}, \bibinfo{author}{Li, C.}, \bibinfo{author}{Xu, C.},
  \bibinfo{year}{2018}.
\newblock \bibinfo{title}{Autoencoder inspired unsupervised feature selection},
  in: \bibinfo{booktitle}{2018 IEEE International Conference on Acoustics,
  Speech and Signal Processing (ICASSP)}, \bibinfo{organization}{IEEE}. pp.
  \bibinfo{pages}{2941--2945}.
\bibitem[{Hashemi et~al.(2019)Hashemi, Ghasemi, Vikalo and Topcu}]{hashemi2019}
\bibinfo{author}{Hashemi, A.}, \bibinfo{author}{Ghasemi, M.},
  \bibinfo{author}{Vikalo, H.}, \bibinfo{author}{Topcu, U.},
  \bibinfo{year}{2019}.
\newblock \bibinfo{title}{Submodular observation selection and information
  gathering for quadratic models}, in: \bibinfo{booktitle}{International
  Conference on Machine Learning}, \bibinfo{organization}{PMLR}. pp.
  \bibinfo{pages}{2653--2662}.
\bibitem[{{Hashemi} et~al.(2020){Hashemi}, {Ghasemi}, {Vikalo} and
  {Topcu}}]{hashemi2020}
\bibinfo{author}{{Hashemi}, A.}, \bibinfo{author}{{Ghasemi}, M.},
  \bibinfo{author}{{Vikalo}, H.}, \bibinfo{author}{{Topcu}, U.},
  \bibinfo{year}{2020}.
\newblock \bibinfo{title}{Randomized greedy sensor selection: Leveraging weak
  submodularity}.
\newblock \bibinfo{journal}{IEEE Transactions on Automatic Control} ,
  \bibinfo{pages}{1--1}.
\bibitem[{Hull(1994)}]{hull1994USPS}
\bibinfo{author}{Hull, J.J.}, \bibinfo{year}{1994}.
\newblock \bibinfo{title}{A database for handwritten text recognition
  research}.
\newblock \bibinfo{journal}{IEEE Transactions on Pattern Analysis and Machine
  Intelligence} \bibinfo{volume}{16}, \bibinfo{pages}{550--554}.
\bibitem[{Iyer et~al.(2013)Iyer, Jegelka and Bilmes}]{iyer2013curvature}
\bibinfo{author}{Iyer, R.K.}, \bibinfo{author}{Jegelka, S.},
  \bibinfo{author}{Bilmes, J.A.}, \bibinfo{year}{2013}.
\newblock \bibinfo{title}{Curvature and optimal algorithms for learning and
  minimizing submodular functions}, in: \bibinfo{booktitle}{Advances in Neural
  Information Processing Systems}, pp. \bibinfo{pages}{2742--2750}.
\bibitem[{Jeffers(1967)}]{Jeffers}
\bibinfo{author}{Jeffers, J.}, \bibinfo{year}{1967}.
\newblock \bibinfo{title}{Two case studies in the application of principal
  component analysis}.
\newblock \bibinfo{journal}{Applied Statistics} , \bibinfo{pages}{225--236}.
\bibitem[{Jolliffe(1986)}]{jolliffe1986principal}
\bibinfo{author}{Jolliffe, I.T.}, \bibinfo{year}{1986}.
\newblock \bibinfo{title}{Principal component analysis and factor analysis},
  in: \bibinfo{booktitle}{Principal Component Analysis}.
  \bibinfo{publisher}{Springer}, pp. \bibinfo{pages}{115--128}.
\bibitem[{Jolliffe et~al.(2003)Jolliffe, Trendafilov and
  Uddin}]{jolliffe2003modified}
\bibinfo{author}{Jolliffe, I.T.}, \bibinfo{author}{Trendafilov, N.T.},
  \bibinfo{author}{Uddin, M.}, \bibinfo{year}{2003}.
\newblock \bibinfo{title}{A modified principal component technique based on the
  {LASSO}}.
\newblock \bibinfo{journal}{Journal of Computational and Graphical Statistics}
  \bibinfo{volume}{12}, \bibinfo{pages}{531--547}.
\bibitem[{Joshi and Boyd(2008)}]{joshi2008sensor}
\bibinfo{author}{Joshi, S.}, \bibinfo{author}{Boyd, S.}, \bibinfo{year}{2008}.
\newblock \bibinfo{title}{Sensor selection via convex optimization}.
\newblock \bibinfo{journal}{IEEE Transactions on Signal Processing}
  \bibinfo{volume}{57}, \bibinfo{pages}{451--462}.
\bibitem[{Kersting et~al.(2007)Kersting, Plagemann, Pfaff and
  Burgard}]{kersting2007most}
\bibinfo{author}{Kersting, K.}, \bibinfo{author}{Plagemann, C.},
  \bibinfo{author}{Pfaff, P.}, \bibinfo{author}{Burgard, W.},
  \bibinfo{year}{2007}.
\newblock \bibinfo{title}{Most likely heteroscedastic gaussian process
  regression}, in: \bibinfo{booktitle}{Proceedings of the 24th International
  Conference on Machine Learning}, \bibinfo{organization}{ACM}. pp.
  \bibinfo{pages}{393--400}.
\bibitem[{Krause et~al.(2008)Krause, Singh and Guestrin}]{krause2008near}
\bibinfo{author}{Krause, A.}, \bibinfo{author}{Singh, A.},
  \bibinfo{author}{Guestrin, C.}, \bibinfo{year}{2008}.
\newblock \bibinfo{title}{Near-optimal sensor placements in gaussian processes:
  Theory, efficient algorithms and empirical studies}.
\newblock \bibinfo{journal}{Journal of Machine Learning Research}
  \bibinfo{volume}{9}, \bibinfo{pages}{235--284}.
\bibitem[{Liu et~al.(2016)Liu, Chepuri, Fardad, Ma{\c{s}}azade, Leus and
  Varshney}]{liu2016sensor}
\bibinfo{author}{Liu, S.}, \bibinfo{author}{Chepuri, S.P.},
  \bibinfo{author}{Fardad, M.}, \bibinfo{author}{Ma{\c{s}}azade, E.},
  \bibinfo{author}{Leus, G.}, \bibinfo{author}{Varshney, P.K.},
  \bibinfo{year}{2016}.
\newblock \bibinfo{title}{Sensor selection for estimation with correlated
  measurement noise}.
\newblock \bibinfo{journal}{IEEE Transactions on Signal Processing}
  \bibinfo{volume}{64}, \bibinfo{pages}{3509--3522}.
\bibitem[{Masaeli et~al.(2010)Masaeli, Yan, Cui, Fung and
  Dy}]{masaeli2010convex}
\bibinfo{author}{Masaeli, M.}, \bibinfo{author}{Yan, Y.}, \bibinfo{author}{Cui,
  Y.}, \bibinfo{author}{Fung, G.}, \bibinfo{author}{Dy, J.G.},
  \bibinfo{year}{2010}.
\newblock \bibinfo{title}{Convex principal feature selection}, in:
  \bibinfo{booktitle}{Proceedings of the 2010 SIAM International Conference on
  Data Mining}, \bibinfo{organization}{SIAM}. pp. \bibinfo{pages}{619--628}.
\bibitem[{McLoone et~al.(2018)McLoone, Johnston and
  Susto}]{mcloone2018methodology}
\bibinfo{author}{McLoone, S.}, \bibinfo{author}{Johnston, A.},
  \bibinfo{author}{Susto, G.A.}, \bibinfo{year}{2018}.
\newblock \bibinfo{title}{A methodology for efficient dynamic spatial sampling
  and reconstruction of wafer profiles}.
\newblock \bibinfo{journal}{IEEE Transactions on Automation Science and
  Engineering} \bibinfo{volume}{15}, \bibinfo{pages}{1692--1703}.
\bibitem[{Minoux(1978)}]{minoux1978}
\bibinfo{author}{Minoux, M.}, \bibinfo{year}{1978}.
\newblock \bibinfo{title}{Accelerated greedy algorithms for maximizing
  submodular set functions}, in: \bibinfo{booktitle}{Optimization techniques}.
  \bibinfo{publisher}{Springer}, pp. \bibinfo{pages}{234--243}.
\bibitem[{Nemhauser et~al.(1978)Nemhauser, Wolsey and
  Fisher}]{nemhauser1978analysis}
\bibinfo{author}{Nemhauser, G.L.}, \bibinfo{author}{Wolsey, L.A.},
  \bibinfo{author}{Fisher, M.L.}, \bibinfo{year}{1978}.
\newblock \bibinfo{title}{An analysis of approximations for maximizing
  submodular set functions—i}.
\newblock \bibinfo{journal}{Mathematical Programming} \bibinfo{volume}{14},
  \bibinfo{pages}{265--294}.
\bibitem[{Olszewski(2001)}]{olszewski2001generalized}
\bibinfo{author}{Olszewski, R.T.}, \bibinfo{year}{2001}.
\newblock \bibinfo{title}{Generalized feature extraction for structural pattern
  recognition in time-series data}.
\newblock \bibinfo{publisher}{PhD thesis, Carnegie Mellon University}.
\bibitem[{Prakash et~al.(2012)Prakash, Honari, Johnston and
  McLoone}]{Prakash2012a}
\bibinfo{author}{Prakash, P.}, \bibinfo{author}{Honari, B.},
  \bibinfo{author}{Johnston, A.}, \bibinfo{author}{McLoone, S.},
  \bibinfo{year}{2012}.
\newblock \bibinfo{title}{Optimal wafer site selection using forward selection
  component analysis}, in: \bibinfo{booktitle}{2012 SEMI Advanced Semiconductor
  Manufacturing Conference}, \bibinfo{organization}{IEEE}. pp.
  \bibinfo{pages}{91--96}.
\bibitem[{Puggini and McLoone(2017)}]{IEEEPuggini}
\bibinfo{author}{Puggini, L.}, \bibinfo{author}{McLoone, S.},
  \bibinfo{year}{2017}.
\newblock \bibinfo{title}{Forward selection component analysis: Algorithms and
  applications}.
\newblock \bibinfo{journal}{IEEE Transactions on Pattern Analysis and Machine
  Intelligence} \bibinfo{volume}{39}, \bibinfo{pages}{2395--2408}.
\bibitem[{Puggini and McLoone(2018)}]{puggini2018_anomaly}
\bibinfo{author}{Puggini, L.}, \bibinfo{author}{McLoone, S.},
  \bibinfo{year}{2018}.
\newblock \bibinfo{title}{An enhanced variable selection and isolation forest
  based methodology for anomaly detection with oes data}.
\newblock \bibinfo{journal}{Engineering Applications of Artificial
  Intelligence} \bibinfo{volume}{67}, \bibinfo{pages}{126--135}.
\bibitem[{Ragnoli et~al.(2009)Ragnoli, McLoone, Lynn, Ringwood and
  Macgearailt}]{ragnoli2009identifying}
\bibinfo{author}{Ragnoli, E.}, \bibinfo{author}{McLoone, S.},
  \bibinfo{author}{Lynn, S.}, \bibinfo{author}{Ringwood, J.},
  \bibinfo{author}{Macgearailt, N.}, \bibinfo{year}{2009}.
\newblock \bibinfo{title}{Identifying key process characteristics and
  predicting etch rate from high-dimension datasets}, in:
  \bibinfo{booktitle}{2009 IEEE/SEMI Advanced Semiconductor Manufacturing
  Conference}, \bibinfo{organization}{IEEE}. pp. \bibinfo{pages}{106--111}.
\bibitem[{Ranieri et~al.(2014)Ranieri, Chebira and Vetterli}]{ranieri2014near}
\bibinfo{author}{Ranieri, J.}, \bibinfo{author}{Chebira, A.},
  \bibinfo{author}{Vetterli, M.}, \bibinfo{year}{2014}.
\newblock \bibinfo{title}{Near-optimal sensor placement for linear inverse
  problems}.
\newblock \bibinfo{journal}{IEEE Transactions on Signal Processing}
  \bibinfo{volume}{62}, \bibinfo{pages}{1135--1146}.
\bibitem[{Rao et~al.(2015)Rao, Chepuri and Leus}]{rao2015greedy}
\bibinfo{author}{Rao, S.}, \bibinfo{author}{Chepuri, S.P.},
  \bibinfo{author}{Leus, G.}, \bibinfo{year}{2015}.
\newblock \bibinfo{title}{Greedy sensor selection for non-linear models}, in:
  \bibinfo{booktitle}{2015 IEEE 6th International Workshop on Computational
  Advances in Multi-Sensor Adaptive Processing (CAMSAP)},
  \bibinfo{organization}{IEEE}. pp. \bibinfo{pages}{241--244}.
\bibitem[{Rodriguez-Lujan et~al.(2014)Rodriguez-Lujan, Fonollosa, Vergara,
  Homer and Huerta}]{rodriguez2014calibration}
\bibinfo{author}{Rodriguez-Lujan, I.}, \bibinfo{author}{Fonollosa, J.},
  \bibinfo{author}{Vergara, A.}, \bibinfo{author}{Homer, M.},
  \bibinfo{author}{Huerta, R.}, \bibinfo{year}{2014}.
\newblock \bibinfo{title}{On the calibration of sensor arrays for pattern
  recognition using the minimal number of experiments}.
\newblock \bibinfo{journal}{Chemometrics and Intelligent Laboratory Systems}
  \bibinfo{volume}{130}, \bibinfo{pages}{123--134}.
\bibitem[{Sun et~al.(2017)Sun, Huang, Wong and Jang}]{sun2017design}
\bibinfo{author}{Sun, K.}, \bibinfo{author}{Huang, S.H.},
  \bibinfo{author}{Wong, D.S.H.}, \bibinfo{author}{Jang, S.S.},
  \bibinfo{year}{2017}.
\newblock \bibinfo{title}{Design and application of a variable selection method
  for multilayer perceptron neural network with {LASSO}}.
\newblock \bibinfo{journal}{IEEE Transactions on Neural Networks and Learning
  Systems} \bibinfo{volume}{28}, \bibinfo{pages}{1386--1396}.
\bibitem[{Susto et~al.(2019)Susto, Maggipinto, Zocco and
  McLoone}]{susto2019induced}
\bibinfo{author}{Susto, G.A.}, \bibinfo{author}{Maggipinto, M.},
  \bibinfo{author}{Zocco, F.}, \bibinfo{author}{McLoone, S.},
  \bibinfo{year}{2019}.
\newblock \bibinfo{title}{Induced start dynamic sampling for wafer metrology
  optimization}.
\newblock \bibinfo{journal}{IEEE Transactions on Automation Science and
  Engineering} \bibinfo{volume}{17}, \bibinfo{pages}{418--432}.
\bibitem[{Sviridenko et~al.(2017)Sviridenko, Vondr{\'a}k and
  Ward}]{sviridenko2017optimal}
\bibinfo{author}{Sviridenko, M.}, \bibinfo{author}{Vondr{\'a}k, J.},
  \bibinfo{author}{Ward, J.}, \bibinfo{year}{2017}.
\newblock \bibinfo{title}{Optimal approximation for submodular and supermodular
  optimization with bounded curvature}.
\newblock \bibinfo{journal}{Mathematics of Operations Research}
  \bibinfo{volume}{42}, \bibinfo{pages}{1197--1218}.
\bibitem[{Tan and San~Lau(2014)}]{tan2014time}
\bibinfo{author}{Tan, S.C.}, \bibinfo{author}{San~Lau, J.P.},
  \bibinfo{year}{2014}.
\newblock \bibinfo{title}{Time series clustering: A superior alternative for
  market basket analysis}, in: \bibinfo{booktitle}{Proceedings of the First
  International Conference on Advanced Data and Information Engineering
  (DaEng-2013)}, \bibinfo{organization}{Springer, Singapore}. pp.
  \bibinfo{pages}{241--248}.
\bibitem[{Tan et~al.(2015)Tan, San~Lau and Yu}]{tan2015finding}
\bibinfo{author}{Tan, S.C.}, \bibinfo{author}{San~Lau, P.},
  \bibinfo{author}{Yu, X.}, \bibinfo{year}{2015}.
\newblock \bibinfo{title}{Finding similar time series in sales transaction
  data}, in: \bibinfo{booktitle}{International Conference on Industrial,
  Engineering and Other Applications of Applied Intelligent Systems},
  \bibinfo{organization}{Springer, Cham}. pp. \bibinfo{pages}{645--654}.
\bibitem[{Van Der~Maaten et~al.(2009)Van Der~Maaten, Postma and Van~den
  Herik}]{Maaten2009}
\bibinfo{author}{Van Der~Maaten, L.}, \bibinfo{author}{Postma, E.},
  \bibinfo{author}{Van~den Herik, J.}, \bibinfo{year}{2009}.
\newblock \bibinfo{title}{Dimensionality reduction: a comparative review}.
\newblock \bibinfo{journal}{Technical Report TiCC-TR 2009-005, Tilburg
  University} .
\bibitem[{Vergara et~al.(2012)Vergara, Vembu, Ayhan, Ryan, Homer and
  Huerta}]{vergara2012chemical}
\bibinfo{author}{Vergara, A.}, \bibinfo{author}{Vembu, S.},
  \bibinfo{author}{Ayhan, T.}, \bibinfo{author}{Ryan, M.A.},
  \bibinfo{author}{Homer, M.L.}, \bibinfo{author}{Huerta, R.},
  \bibinfo{year}{2012}.
\newblock \bibinfo{title}{Chemical gas sensor drift compensation using
  classifier ensembles}.
\newblock \bibinfo{journal}{Sensors and Actuators B: Chemical}
  \bibinfo{volume}{166}, \bibinfo{pages}{320--329}.
\bibitem[{Waleesuksan and Wongsa(2016)}]{waleesuksan2016fast}
\bibinfo{author}{Waleesuksan, C.}, \bibinfo{author}{Wongsa, S.},
  \bibinfo{year}{2016}.
\newblock \bibinfo{title}{A fast variable selection for nonnegative
  garrote-based artificial neural network}, in: \bibinfo{booktitle}{13th
  International Conference on Electrical Engineering/Electronics, Computer,
  Telecommunications and Information Technology (ECTI-CON)},
  \bibinfo{organization}{IEEE}. pp. \bibinfo{pages}{1--6}.
\bibitem[{Wang et~al.(2016)Wang, Moran, Wang and Pan}]{wang2016}
\bibinfo{author}{Wang, Z.}, \bibinfo{author}{Moran, B.}, \bibinfo{author}{Wang,
  X.}, \bibinfo{author}{Pan, Q.}, \bibinfo{year}{2016}.
\newblock \bibinfo{title}{Approximation for maximizing monotone non-decreasing
  set functions with a greedy method}.
\newblock \bibinfo{journal}{Journal of Combinatorial Optimization}
  \bibinfo{volume}{31}, \bibinfo{pages}{29--43}.
\bibitem[{Wei and Billings(2007)}]{wei2007feature}
\bibinfo{author}{Wei, H.L.}, \bibinfo{author}{Billings, S.A.},
  \bibinfo{year}{2007}.
\newblock \bibinfo{title}{Feature subset selection and ranking for data
  dimensionality reduction}.
\newblock \bibinfo{journal}{IEEE Transactions on Pattern Analysis and Machine
  Intelligence} \bibinfo{volume}{29}.
\bibitem[{Whitley et~al.(2000)Whitley, Ford and
  Livingstone}]{whitley2000unsupervised}
\bibinfo{author}{Whitley, D.C.}, \bibinfo{author}{Ford, M.G.},
  \bibinfo{author}{Livingstone, D.J.}, \bibinfo{year}{2000}.
\newblock \bibinfo{title}{Unsupervised forward selection: a method for
  eliminating redundant variables}.
\newblock \bibinfo{journal}{Journal of Chemical Information and Computer
  Sciences} \bibinfo{volume}{40}, \bibinfo{pages}{1160--1168}.
\bibitem[{Witten et~al.(2009)Witten, Tibshirani and
  Hastie}]{witten2009penalized}
\bibinfo{author}{Witten, D.M.}, \bibinfo{author}{Tibshirani, R.},
  \bibinfo{author}{Hastie, T.}, \bibinfo{year}{2009}.
\newblock \bibinfo{title}{A penalized matrix decomposition, with applications
  to sparse principal components and canonical correlation analysis}.
\newblock \bibinfo{journal}{Biostatistics} \bibinfo{volume}{10},
  \bibinfo{pages}{515--534}.
\bibitem[{Wold(1973)}]{wold1973nonlinear}
\bibinfo{author}{Wold, H.}, \bibinfo{year}{1973}.
\newblock \bibinfo{title}{Nonlinear iterative partial least squares ({NIPALS})
  modelling: Some current developments}, in: \bibinfo{booktitle}{Multivariate
  Analysis--III}. \bibinfo{publisher}{Elsevier}, pp. \bibinfo{pages}{383--407}.
\bibitem[{Zhou et~al.(2014)Zhou, Claire and King}]{zhou2014predicting}
\bibinfo{author}{Zhou, F.}, \bibinfo{author}{Claire, Q.},
  \bibinfo{author}{King, R.D.}, \bibinfo{year}{2014}.
\newblock \bibinfo{title}{Predicting the geographical origin of music}, in:
  \bibinfo{booktitle}{2014 IEEE International Conference on Data Mining
  (ICDM)}, \bibinfo{organization}{IEEE}. pp. \bibinfo{pages}{1115--1120}.
\bibitem[{Zocco and McLoone(2017)}]{zocco2017mean}
\bibinfo{author}{Zocco, F.}, \bibinfo{author}{McLoone, S.},
  \bibinfo{year}{2017}.
\newblock \bibinfo{title}{Mean squared error vs. frame potential for
  unsupervised variable selection}, in: \bibinfo{booktitle}{Intelligent
  Computing, Networked Control, and Their Engineering Applications}.
  \bibinfo{publisher}{Springer}, pp. \bibinfo{pages}{353--362}.
\bibitem[{Zou et~al.(2006)Zou, Hastie and Tibshirani}]{zou2006sparse}
\bibinfo{author}{Zou, H.}, \bibinfo{author}{Hastie, T.},
  \bibinfo{author}{Tibshirani, R.}, \bibinfo{year}{2006}.
\newblock \bibinfo{title}{Sparse principal component analysis}.
\newblock \bibinfo{journal}{Journal of Computational and Graphical Statistics}
  \bibinfo{volume}{15}, \bibinfo{pages}{265--286}.

\end{thebibliography}

\end{document}